\documentclass{article} % For LaTeX2e
\usepackage{iclr2026_conference,times}

\usepackage{amsmath,amsfonts,bm}

\def\eqref#1{equation~\ref{#1}}
\def\1{\bm{1}}

\DeclareMathAlphabet{\mathsfit}{\encodingdefault}{\sfdefault}{m}{sl}
\SetMathAlphabet{\mathsfit}{bold}{\encodingdefault}{\sfdefault}{bx}{n}

\usepackage{hyperref}
\usepackage{url}

\usepackage[table]{xcolor}
\usepackage{listings}
\usepackage{epsfig}
\usepackage{amsmath}
\usepackage{amssymb}
\usepackage{verbatim}
\usepackage{booktabs}
\usepackage{amsmath}
\usepackage{tabularx}
\usepackage{amsfonts,amssymb}
\usepackage{multirow}
\usepackage{bbding}
\usepackage{algorithm,algorithmicx,algpseudocode}
\usepackage{array}
\usepackage{colortbl}
\usepackage{natbib}
\usepackage{wasysym}
\usepackage{wrapfig}
\usepackage{tabularx}
\usepackage{markdown}
\usepackage{subcaption}

\definecolor{mygray}{gray}{.92}
\definecolor{lightgray}{gray}{.96}
\definecolor{myy}{RGB}{126,95,0}
\definecolor{ggray}{RGB}{127,127,127}
\definecolor{mygreen}{RGB}{0,0,0}
\definecolor{myred}{RGB}{240,16,89}
\definecolor{myblue}{RGB}{0,114,188}
\definecolor{darkgreen}{rgb}{0.0, 0.5, 0.0}
\definecolor{demphcolor}{RGB}{100,100,100}

\makeatletter
\newcommand{\thickhline}{%
    \noalign {\ifnum 0=`}\fi \hrule height 0.8pt
    \futurelet \reserved@a \@xhline
}

\usepackage{etoolbox}
\makeatletter
\AfterEndEnvironment{algorithm}{\let\@algcomment\relax}
\AtEndEnvironment{algorithm}{\kern2pt\hrule\relax\vskip3pt\@algcomment}
\let\@algcomment\relax
\newcommand\algcomment[1]{\def\@algcomment{\footnotesize#1}}
\renewcommand\fs@ruled{\def\@fs@cfont{\bfseries}\let\@fs@capt\floatc@ruled
  \def\@fs@pre{\hrule height.8pt depth0pt \kern2pt}%
  \def\@fs@post{}%
  \def\@fs@mid{\kern2pt\hrule\kern2pt}%
  \let\@fs@iftopcapt\iftrue}
\makeatother
\definecolor{mygray}{gray}{.92}
\definecolor{mygreen}{RGB}{93,173,85}

\makeatother

\newcolumntype{d}[1]{>{\raggedright\arraybackslash}p{#1pt}}
\newcolumntype{e}[1]{>{\raggedleft\arraybackslash}p{#1pt}}

\definecolor{baselinecolor}{gray}{.9}

\newlength\savewidth

\renewcommand{\paragraph}[1]{\vspace{1.25mm}\noindent\textbf{#1}}

\newcolumntype{x}[1]{>{\centering\arraybackslash}p{#1pt}}
\newcolumntype{y}[1]{>{\raggedright\arraybackslash}p{#1pt}}
\newcolumntype{z}[1]{>{\raggedleft\arraybackslash}p{#1pt}}

\newcommand{\app}{\raise.17ex\hbox{$\scriptstyle\sim$}}

\definecolor{deemph}{gray}{0.6}

\definecolor{baselinecolor}{gray}{.9}

\definecolor{color_green}{HTML}{92D050}

\makeatletter
\patchcmd{\ALG@name}{Algorithm}{\textbf{Algorithm}}{}{}
\makeatother

\title{Discrete Diffusion Models with MLLMs for Unified Medical Multimodal Generation}
\author{%
\bf \textbf{Jiawei Mao}$^{1}$
\quad
\textbf{Yuhan Wang}$^{1}$ \quad
\textbf{Lifeng Chen}$^{1 }$ \quad 
\textbf{Can Zhao}$^2$ \quad
\bf \textbf{Yucheng Tang}$^2$ \\
\textbf{Dong Yang}$^2$ \quad
\textbf{Liangqiong Qu}$^3$ \quad
\textbf{Daguang Xu}$^2$ \quad
\textbf{Yuyin Zhou}$^1$ \vspace{.3em} \\
$^1$UC Santa Cruz
\quad $^2$NVIDIA \quad $^3$University of Hong Kong\\
\small \includegraphics[height=1.1em]{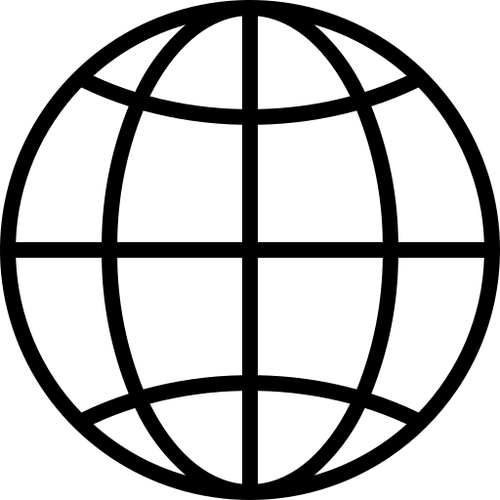} \textbf{Project Page}: \url{https://jwmao1.github.io/MeDiM_web/} \\
\small
\includegraphics[height=1.2em]{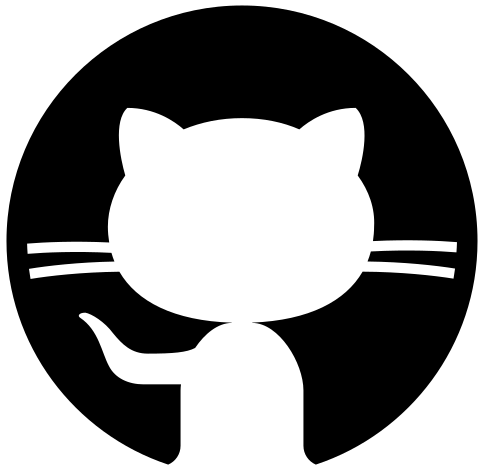} \textbf{Model Training}: \url{https://github.com/UCSC-VLAA/MeDiM} \\
}

\iclrfinalcopy % Uncomment for camera-ready version, but NOT for submission.
\begin{document}

\maketitle

\begin{figure}[h!]
    \centering
    \vspace{-0.3cm}
    \includegraphics[width=0.95\linewidth]{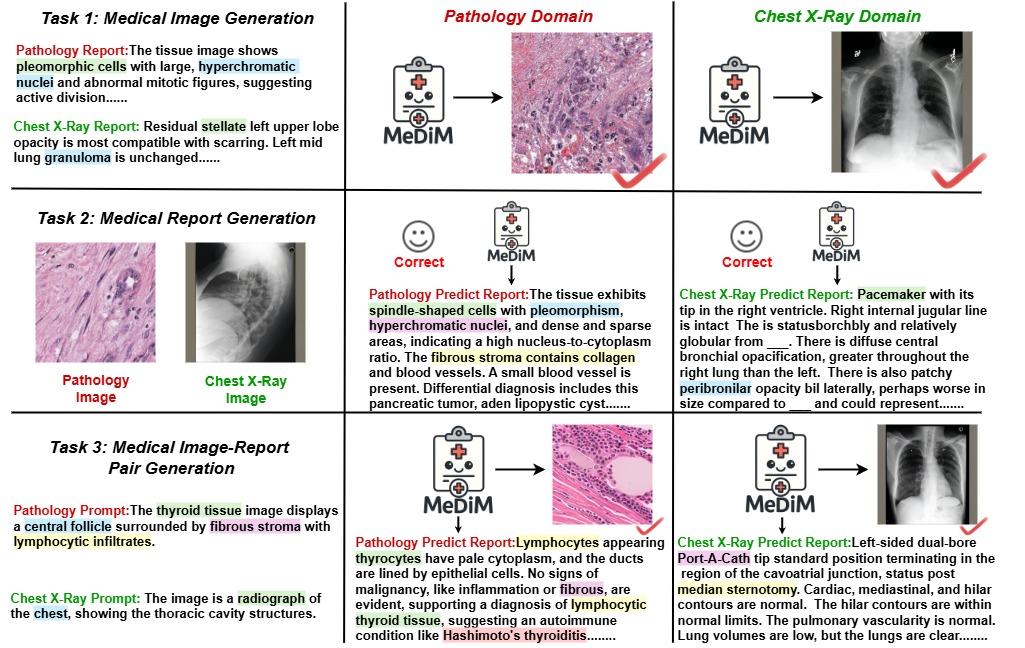}
    \vspace{-0.3cm}
     \caption{
     \textbf{MeDiM}, the \emph{first \textbf{medical discrete diffusion model}}, is a flexible multimodal generator that simultaneously supports: \textbf{(i)} medical image generation from clinical reports, \textbf{(ii)} report generation from medical images, and \textbf{(iii)} joint synthesis of image–report pairs.
     Zoom in for a better view.
     }
     \label{fig_teaser}
   \end{figure}

\begin{abstract}
% General medical image analysis (i.e., understanding and generation) are often limited by data scarcity, incomplete modalities, and inter-modality differences arising from privacy constraints, high annotation costs, and organ-specific variations. 
% This paper presents \textbf{MeDiM}, a new class of medical diffusion models that provide unified solutions for diverse medical tasks in multiple medical modalities, including image generation, report generation, and paired image-report synthesis. 
% The key is through discrete diffusion modeling, unifying the modeling of shared probabilistic distributions across vision and language modalities, removing the need for modality-specific components and enables the generation of semantically consistent paired data. 
% To further extend into a flexible tool capable of multimodal medcial iamge anlsysis We adapt multimodal large language models (MLLMs) as interface of the diffusion backbone, allowing finegrained alignment between textual and visual modalities during generation. to improve cross-modal consistency and produces more coherent and reliable image-report pairs.  
% Extensive experiments show that MeDiM can serve as a unified foundation model for both medical generation and understanding tasks.
% The jointly generated medical pairs improve downstream performance, helping to alleviate data scarcity and annotation costs. 
% Collectively, the results indicate that MLLMs, owing to their intrinsic multimodal reasoning and alignment strengths, are especially suited as backbones for medical paired data generation.

Recent advances in generative medical models are often constrained by modality-specific scenarios that hinder the integration of complementary evidence, such as imaging, pathology, and clinical notes. This fragmentation limits their development to true foundation models that empower medical AI agents to learn from and predict across the full spectrum of biomedical knowledge. To address these challenges, we propose \textbf{MeDiM}, the first medical discrete diffusion model that learns shared distributions across different medical modalities without requiring modality-specific components. MeDiM unifies multiple generative tasks: it flexibly translates between images and text or jointly produces image–report pairs across domains in response to user prompts.
% To address there challenges, we introduces \textbf{MeDiM}, a new class of medical multimodal diffusion model that unifies various generative tasks: it flexibly translates between images and text and jointly generate complete image-report pairs across different domains based on user prompt.
% MeDiM is the first medical discrete diffusion model that models shared distributions in different medical modalities, without requiring modality-specific components.
It builds on a discrete diffusion framework that unifies vision and language representations by modeling their shared probabilistic distribution. To empower the diffusion process to support unified and versatile medical generation, we employ a multimodal large language model (MLLM) as the diffusion backbone, leveraging its rich prior knowledge and cross-modal reasoning abilities. Because MLLMs are trained with causal (autoregressive) masking while diffusion denoising benefits from bidirectional context, MeDiM introduces two key designs: 1) \emph{removing the causal attention mask} to enable a fully bidirectional information flow essential for mutual alignment, and 2) \emph{injecting continuous timestep embeddings} to make the MLLM aware of the diffusion steps. 
% Extensive experiments show that MeDiM can serve as a unified foundation model for a range of medical generation tasks on various domains not only generate high-quality data (16.60 FID and 24.19 FID in medical image generation; 0.265 METEOR scores and 0.258 METEOR scores on medical report generation for MIMIC-CXR and PathGen dataset).
Extensive experiments validate MeDiM as a unified foundation model capable of high-fidelity medical generation across various modalities, including medical image generation (16.60 FID on MIMIC-CXR; 24.19 FID on PathGen) and report generation (0.2650 METEOR on MIMIC-CXR; 0.2580 METEOR on PathGen). In addition, the jointly generated medical image-report pairs improve the downstream task performance (+6.43$\%$ BLEU-1, +18.57$\%$ BLEU-2, +31.58$\%$ BLEU-3, and +4.80$\%$ METEOR in PathGen), enabling the use of multimodal inputs and the production of coherent, clinically grounded outputs. 
\end{abstract}  
\section{Introduction}

% Medical image analysis and medical image generation play an increasingly important role in modern healthcare, supporting disease diagnosis, treatment planning, and clinical decision-making. 

%Recent advances in medical generative models have substantially accelerated progress in these areas by enabling a more accurate understanding of complex medical images~\citep{liu2025enhanced,wang2024hergen,liu2024structural,liu2024context,liu2024bootstrapping} and by facilitating the synthesis of realistic medical images~\citep{guo2025med,li2025u,guo2025maisi,polamreddy2025leapfrog}. 
%However, the advancement of general medical image analysis and generation remains constrained by three major challenges: \textbf{scarce data} due to strict privacy regulations, the \textbf{high cost of expert annotations}, and significant \textbf{inter-modality variability} arising from organ-specific characteristics and differences in imaging devices.

Modern medical systems and doctors rely on synthesizing multimodal evidence, encompassing radiology images, digital pathology images, EHR info and clinical reports. However, most existing medical AI models remain limited to isolated modalities~\citep{moor2023foundation}. 
%This fragmentation is a significant factor in $\sim$12M diagnostic errors in the US annually~\citep{singh2014frequency}, as critical cross-modal correlations go unprocessed. 
They often face limited insights when interpreting complex cases. For instance, current AI tools are unable to jointly analyze a lung nodule's imaging with its corresponding biopsy mutation status to predict treatment resistance, or generate clinically grounded images (e.g., counterfactual follow-up radiological scan or representative pathology patches) that visualize likely outcomes under different therapies. Bridging this gap demands a foundational shift: a unified, domain-aware multimodal system capable of understanding heterogeneous inputs, while generating clinically meaningful outputs. Such a system would directly address the challenge of cross-modal alignment in the medical context and serve as a foundation for medical AI agents that can learn from and generate across the full spectrum of biomedical knowledge.

Medical multimodal synthesis can represent a promising direction toward generalist medical AI agents, but existing methods remain limited. 
% PairAug~\citep{xie2024pairaug} uses GPT-3.5~\citep{brown2020language} and Stable Diffusion~\citep{rombach2022high} to augment medical image–report pairs via inter-patient and intra-patient branches, but its reliance on external models rather than learning of the joint image–report distribution prevents strict semantic alignment. 
% MedM2G~\citep{zhan2024medm2g} adopts a text-centric framework to align multiple modalities and generation in the latent space via cross-guided, with modality-specific components making it difficult to scale to broader medical modalities.
Medical-specific models like PairAug~\citep{xie2024pairaug} and MedM2G~\citep{zhan2024medm2g} either rely on disconnected external models, preventing strict semantic alignment, or use modality-specific components that are difficult to adapt to multiply modality. 
In contrast, the natural image domain has witnessed the emergence of unified models~\citep{team2024chameleon,xie2024show,wu2024liquid,yang2025mmada} that, within a single framework, simultaneously support both generation and understanding tasks without the need for modality-specific designs.
Liquid~\citep{wu2024liquid} extends a pre-trained large language model (LLM) into a unified multimodal auto-regressive (AR) framework, allowing images and text to share a token space for both visual understanding and generation without altering the LLM architecture. 
\citet{swerdlow2025unified} note that while AR models excel in text, their token-by-token prediction limits efficiency, motivating a unified multimodal discrete diffusion model that enables higher-quality, diverse, and controllable generation. 
\citet{yang2025mmada} further propose MMaDA, employing a unified diffusion architecture to jointly model image and text distributions. However, it does not support the paired generation of image-text outputs, a critical capability needed to address the medical challenges outlined above.

\begin{figure*}[!t]
    \centering
    \setlength{\abovecaptionskip}{0mm} %调整caption与图的距离
    \setlength{\belowcaptionskip}{-5mm}%调整caption与下文的距离
    \includegraphics[width=1\linewidth]{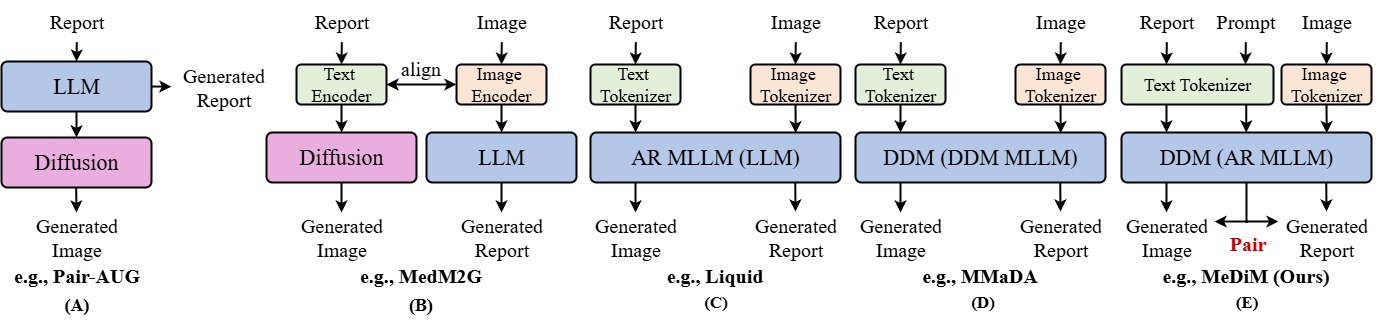}
     \caption{\textbf{Architectural comparison of medical multimodal models.} \textbf{(“BACKBONE'')} indicates the backbone adopted in each framework. Prior approaches (A-D) cannot perform paired generation and suffer from other key limitations, such as requiring modality-specific components (A, B), inference inefficiency (C), or backbone inflexibility (D). In contrast, our model, MeDiM (E), provides a unified framework designed to overcome these challenges.}
     % However, (A), (B), (C), and (D) are all incapable of supporting multimodal paired generation. Moreover, (A) and (B) require modality-specific components. Although (C) and (D) do not require additional designs for different modalities, (C) suffers from low inference efficiency, while (D) is constrained by a single backbone that lacks generality and flexibility. In contrast, MeDiM (E) is better suited as a framework for medical multimodal paired generation.}
     \label{fig_paradigm}
   \end{figure*}

To our knowledge, no such unified models currently exist in the medical domain that could synthesize multimodal information while supporting multimodal generation (see Fig.~\ref{fig_paradigm}). 
In this work, we propose \textbf{MeDiM}, the first medical discrete diffusion model that simultaneously models shared distributions across different modalities. 
Compared to domain-specific expert models, MeDiM can simultaneously perform diverse medical tasks across multiple medical modalities and domains, including medical image/report generation, and medical paired image–report synthesis (as shown in Fig.~\ref{fig_teaser}). 
% We investigate the architectural design of discrete diffusion models (DDM) for medical multimodal paired generation and provide empirical insights for future research on modeling shared distributions in the medical domain. 
A core design is the use of a Multimodal Large Language Model (MLLM) as the backbone for the diffusion process. Pre-trained MLLMs provide strong distribution-alignment priors from large-scale vision–language pretraining, making them powerful guides for multimodal generation. Their increasingly unified architectures are particularly well suited for paired image–report generation. Unlike MMaDA~\citep{yang2025mmada}, which is restricted to diffusion-based MLLM backbones~\citep{nie2025large}, our MeDiM can extend to a broader class of autoregressive (AR) MLLMs, offering greater generality and flexibility.
However, adapting MLLMs to discrete diffusion introduces a fundamental mismatch: MLLMs are trained with a causal (autoregressive) attention mask, while the multimodal diffusion denoising process is inherently non-causal (e.g., requires bidirectional context).  To resolve this, MeDiM incorporates two key modifications: (1) \textbf{causal attention removal}, enabling full bidirectional information flow for improved cross-modal alignment, and (2) \textbf{injecting continuous timestep embeddings}, allowing the MLLM to track diffusion steps. In addition, we integrate adaptive layer normalization (AdaLN)~\citep{perez2018film,brock2018large,karras2019style} to further stabilize the training and enhance its generative capability.

% This design is motivated by pre-trained multimodal large language models (MLLMs), with their strong distribution-alignment priors from large-scale vision–language pretraining, serve as powerful backbones for guiding multimodal generation. Recent MLLM architectures adopt increasingly unified designs, making them particularly suitable for paired image–report generation. In contrast to MMaDA~\citep{yang2025mmada}, which is tied to diffusion-based MLLM backbones~\citep{nie2025large}, MeDiM extends to a broader class of AR-based MLLMs, offering greater generality and flexibility. Because MLLMs are trained with causal (autoregressive) masking while disc diffusion denoising benefits from bidirectional context, MeDiM introduces two key designs: 1) \emph{removing the causal attention mask} to enable a fully bidirectional information flow essential for mutual alignment, and 2) \emph{injecting continuous timestep embeddings} to make the MLLM aware of the diffusion steps. To adapt MLLMs to discrete diffusion, we further incorporate architectural refinements, including causal mask removal, timestep embeddings, and adaptive layer normalization (AdaLN)~\citep{perez2018film,brock2018large,karras2019style}.

Our experiments demonstrate that MeDiM can function as a versatile foundation model for unifying various medical generative tasks: 
1) For medical image generation, MeDiM achieves \underline{\emph{state-of-the-art Frechet Inception Distances (FID)}} on the MIMIC-CXR~\citep{johnson2019mimic} and PathGen~\citep{sun2024pathgen} datasets, respectively (Tab.~\ref{tab_1} and Tab.~\ref{tab_2}; 16.60 FID and 24.19 FID), generating \underline{\emph{high-fidelity}} medical images across \underline{\emph{different modalities}} (Fig.~\ref{fig_qual}; robust and high visual quality). 
2) For medical report generation, MeDiM generates corresponding clinical reports from input medical images, demonstrating \underline{\emph{semantic alignment}} with target reports in MIMIC-CXR and PathGen datasets, respectively (Tab.~\ref{tab_report}; METEOR score of 0.265 and 0.258). 
3) MeDiM generates \underline{\emph{highly consistent}} (Fig.~\ref{fig_joint_a}; higher consistency score in both large vision–language models (VLM) and human evaluation) medical image–report pairs. The generated medical image–report pairs can further \underline{\emph{improve the performance of VLM}} on downstream medical report generation tasks (Fig.~\ref{fig_joint_c}; +6.43$\%$ BLEU-1, +18.57$\%$ BLEU-2, +31.58$\%$ BLEU-3, and +4.80$\%$ METEOR in PathGen). 
4) Comparative \underline{\emph{analyses of backbone choices}} (Sec.~\ref{sec_backbone}; improve with MLLM backbone) reveal that MLLM backbones are particularly well-suited for multimodal paired generation in medical discrete diffusion models.

In summary, the key contributions of this work can be distinguished in the following aspects:

$\bullet$ We propose MeDiM, the first medical discrete diffusion model that models shared distributions in different medical modalities, without requiring modality-specific components.

$\bullet$ MLLMs with distribution-alignment priors are identified as superior backbones for discrete diffusion models in medical multimodal generation.

$\bullet$ MeDiM demonstrates state-of-the-art or competitive performance in unified medical image analysis and generation tasks, with the generated medical image–report pairs improving the performance of medical vision–language models (VLMs).

\section{Method}
\begin{figure*}[!t]
    \centering
    \setlength{\abovecaptionskip}{2mm} %调整caption与图的距离
    \setlength{\belowcaptionskip}{-5mm}%调整caption与下文的距离
    \includegraphics[width=1\linewidth]{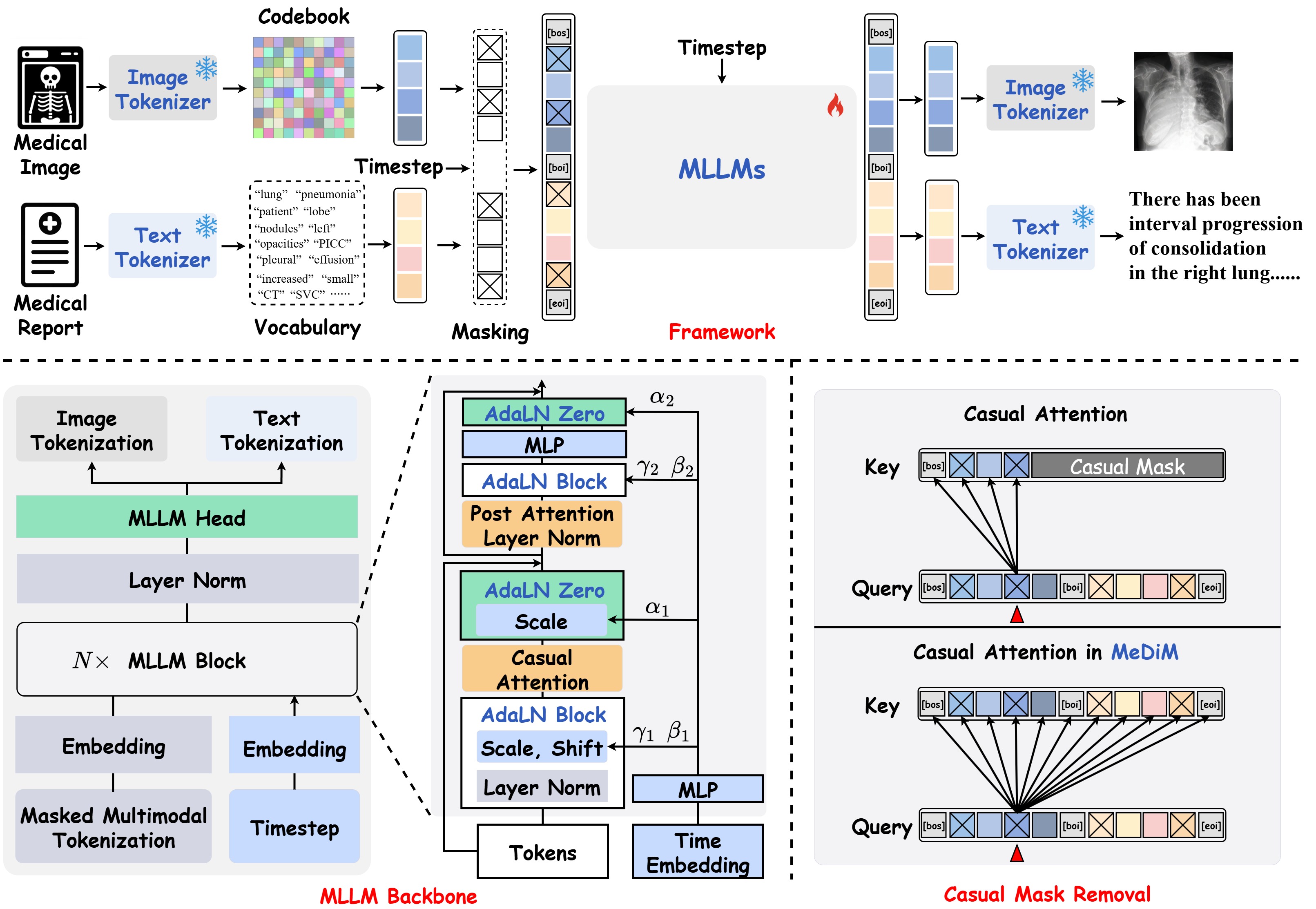}
     \caption{\textbf{Overview of the MeDiM architecture.} The framework integrates an MLLM backbone within a discrete diffusion process for unified medical multimodal generation. During the forward process, data is tokenized and diffused over timesteps. The MLLM is then trained to reverse this process. Key architectural adaptations, including causal attention removal, timestep embeddings, and AdaLN, adapt the autoregressive MLLM for the bidirectional denoising required for unified medical generation.}
     % Illustration of the MeDiM framework and backbone design. The framework integrates a discrete diffusion process with an MLLM backbone, enabling unified modeling of medical images and reports. MeDiM adopts a discrete diffusion framework to integrate medical images and texts into a shared embedding space, where multimodal embeddings are absorbed according to timesteps. The MLLM then leverages temporal awareness to align the masked multimodal embeddings within the unified representation space and predict their absorbed states. To enable the MLLM backbone to fit within the discrete diffusion framework while incorporating temporal awareness, we propose a set of architectural refinements, such as timestep embeddings, AdaLN designs, and the removal of the causal mask.}
     \label{fig_framework}
   \end{figure*}

We propose MeDiM, the first  discrete diffusion model with MLLM designed for medical multimodal generation. 
The framework aims to jointly model the shared distributions between medical images and reports without requiring additional modality-specific components, thereby providing flexibility to accommodate more medical modalities.  
Fig.~\ref{fig_framework} illustrates the overall framework of MeDiM. 
This section introduces discrete diffusion models in the medical domain and MeDiM's architectural design. Related background can be found in Sec.~\ref{sec_related} of appendices.

\subsection{Discrete Diffusion Models}

Diffusion models~\citep{ho2020denoising,rombach2022high} are probabilistic generative models that learn to approximate data distributions by sequentially corrupting and denoising samples.
The forward diffusion process gradually perturbs the data sample $x_{0}$ with noise over a sequence of timesteps $t$, producing a latent distribution $q(x_{t})$:
\begin{equation}
x_t \sim q(x_t \mid x_0) = \mathcal{N}\big(x_t; \sqrt{\bar{\alpha}_t} x_0, (1-\bar{\alpha}_t)\mathbf{I} \big),
\label{ddpm_forward}
\end{equation}
where $\bar{\alpha}_t = \prod_{s=1}^{t} (1-\beta_s)$ denotes the cumulative retention coefficient and $\beta_t$ represents the pre-defined noise variance at timestep $t$.
The reverse diffusion process involves learning a parameterized denoising model $\epsilon(.)$ to iteratively reconstruct the original data from noisy inputs:
\begin{equation}
x_{t-1} = \frac{1}{\sqrt{\alpha_t}} 
\left( 
x_t - \frac{1-\alpha_t}{\sqrt{1-\bar{\alpha}_t}} \, \epsilon(x_t,t) 
\right) 
+ \sqrt{\frac{1-\bar{\alpha}_{t-1}}{1-\bar{\alpha}_t} \, \beta_t} \, z, 
\quad z \sim \mathcal{N}(0, I)
\label{ddpm_backward}
\end{equation}

%However, diffusion models are limited in tasks involving heterogeneous modalities and discrete representations. 
%In medical image–report generation, this limitation is critical, as it requires modeling both visual and textual sequences within a unified framework. 
We introduce a discrete diffusion model~\citep{sohl2015deep} for medical multimodal generation that jointly models medical images and reports within a shared probabilistic space. 

\paragraph{Forward Diffusion.}
Our medical discrete diffusion models operate directly on sequences of discrete symbols $x_0$ consisting of report tokenizations $x_{r0}$ and quantized medical image tokens $x_{i0}$ encoded by VQ-VAE~\citep{van2017neural}. 
The forward diffusion process is formulated as a Markov chain $q(x_t \mid x_{t-1})$, 
where original symbols are gradually replaced with noise symbols, 
until the distribution converges to an approximate uniform distribution at a large timestep $T$. 
This process is parameterized by a transition matrix $Q_t \in \mathbb{R}^{K \times K}$, 
where $K$ denotes the sum of vocabulary size for the text tokenizer and VQ-VAE codebook. 
The elements of $Q_t$ represent the transition probabilities between discrete states, i.e., $[Q_t]_{ij}$ is the probability of transitioning from state $i$ at timestep $t-1$ to state $j$ at timestep $t$.
Considering that absorbing transition matrices yield better performance in multimodal tasks~\citep{austin2021structured,lou2023discrete}, MeDiM introduces a special [MASK] token as an absorbing state, which serves as the noise symbol during the forward diffusion process. 
Please note that with the introduction of the additional [MASK] token, the dimension of the transition matrix is expanded to $Q_t \in \mathbb{R}^{(K+1) \times (K+1)}$.
Accordingly, the transition matrix $Q_t$ for the absorbing state formulation can be expressed as:
\begin{equation}
Q_t = \alpha_t I + (1 - \alpha_t)\,\mathbf{1}\,e_m^\top,
\label{q_distribution}
\end{equation}
here, $\alpha_t \in [0,1]$ is the retention probability, 
$\mathbf{1}$ is a vector filled with ones, and $e_m$ indicates the canonical basis vector that activates only the absorbing [MASK] state $m$. 
This construction ensures that once a token is replaced by the [MASK] symbol, it remains in that state for all subsequent timesteps. 
Consequently, the forward transition distribution is given by a categorical distribution:
\begin{equation}
\begin{split}
& x_0 = [x_{r0},x_{i0}]\\
& q(x_t \mid x_0) = \text{Cat}(x_t; p = \bar{Q_t} x_0),
\label{dsm_diffusion}
\end{split}
\end{equation}
where $\bar{Q}_t = \prod_{s=1}^t Q_s$. 
This implies that, as the number of timesteps increases, the input sequence is progressively replaced by [MASK] tokens, and at a sufficiently large timestep, the distribution converges to the absorbing state, providing a well-defined initialization for the reverse diffusion process. 

\paragraph{Reverse Diffusion.}
The reverse diffusion process aims to reconstruct the original data sequence from noisy inputs by progressively recovering masked or transitioned symbols.
Specifically, given a corrupted sequence $x_t$ at timestep $t$, the discrete diffusion model parameterizes the reverse transition distribution $p_\theta(x_{t-1} \mid x_t)$, which estimates the probability of recovering the clean symbol at the previous step.
Formally, this process is defined as a categorical distribution over the shared vocabulary space:
\begin{equation}
p_\theta(x_{t-1} \mid x_t) = \text{Cat}\big(x_{t-1}; \epsilon(x_t, t)\big),
\label{reverse_discrete}
\end{equation}
where $\epsilon(x_t, t) \in \Delta^{K+1}$ denotes the predicted categorical probabilities at timestep $t$, and $\Delta^{K+1}$ is the probability simplex over the extended vocabulary including the [MASK] token.

To effectively model multimodal medical data, we parameterize $\epsilon(x_t, t)$ using a backbone network $f_\theta(\cdot)$ built upon an MLLM.
MeDiM incorporates timestep embeddings into the MLLM backbone, ensuring temporal conditioning across the transition trajectory.
At each step, $f_\theta$ leverages prior alignment between medical image tokens and report tokens.
Consequently, the reverse chain iteratively recovers $x_0 = [x_{r0}, x_{i0}]$ from a fully masked initialization, producing coherent paired outputs that align medical visual and textual modalities.
The training objective~\citep{sahoo2024simple} can be defined as the expected negative log-likelihood of recovering the original data sequence, weighted by the transition schedule:
\begin{equation}
\mathcal{L} = -\mathbb{E}_{t \sim \mathcal{U}(0,1), q(x_t \mid x)} \left[ \frac{\alpha_t'}{1 - \alpha_t} \log p_\theta(x_0 \mid x_t) \right],
\end{equation}
$\alpha_t' = \alpha_t - \alpha_{t-1}$, i.e., the incremental change in retention probability with respect to timestep $t$.
This objective ensures that the model learns to accurately approximate the reverse transition distribution across the entire transition trajectory.

\subsection{Discrete Diffusion with MLLMs}

% To more effectively align medical reports with medical images during the generation process, we have discussed the backbone design of discrete diffusion models for medical multimodal paired generation. 
% Motivated by the architectural requirements of multimodal generation, we naturally introduce MLLM as the backbone of the discrete diffusion framework.   
% MLLMs provide strong cross-modal alignment priors learned from large-scale vision–language pretraining, which are crucial for ensuring semantic and visual consistency in medical image–report generation. 
% Moreover, their unified token-based representation offers scalability to accommodate diverse medical modalities, such as chest X-rays, CT, MRI, and pathology images. 
% Empirically, Our preliminary results also show that discrete diffusion models for medical multimodal pair generation benefit more from MLLM backbones (see Sec~\ref{sec_backbone} in Supplementary), which exhibit an increasingly unified architecture in recent designs, enabling more consistent medical paired outputs.
% Compared to MMaDA~\citep{yang2025mmada}, which is limited to diffusion-based MLLM backbones~\citep{nie2025large}, MeDiM adopts a more general AR MLLM, thus endowing MeDiM with greater extensibility.
% To adapt MLLMs for discrete diffusion, we introduce causal mask removal, timestep embeddings, and AdaLN designs as detailed below. 

A central design in our framework is the integration of a MLLM as the backbone to empower the discrete diffusion process to support unified medical multimodal generation. This selection is motivated by several key advantages of MLLMs: 1) MLLMs provide powerful cross-modal alignment priors from large-scale vision-language pre-training, which are crucial for ensuring the semantic and visual consistency of generated image-report pairs. 2) Their unified token-based representation offers inherent scalability, allowing the framework to accommodate diverse medical modalities—such as chest X-rays, CT scans, and pathology images—with minimal architectural changes.
Our empirical results further validate this choice, demonstrating that MLLM backbones significantly outperform strong alternatives (e.g., DiT~\cite{Peebles2022DiT})for medical report generation and image-report pair generation tasks (see Sec.~\ref{sec_backbone} in Supplementary). Furthermore, our framework offers greater flexibility than prior MLLM-based diffusion models. While approaches like MMaDA~\citep{yang2025mmada} are limited to diffusion-specific backbones~\citep{nie2025large}, our method is also compatible with a broader class of general-purpose auto-regressive (AR) MLLMs.

Since MLLMs are trained with causal (autoregressive) masking, while diffusion denoising relies on bidirectional context, MeDiM introduces two key adaptations: \emph{causal mask removal} and \emph{timestep embeddings}. In addition, we identify AdaLN as a critical normalization strategy. Together, these three components ensure a seamless integration of MLLMs into the discrete diffusion process, as detailed below.

\paragraph{Causal Mask Removal.}
Autoregressive MLLMs typically employ causal attention, which restricts each token to attend only to its previous context.
While this constraint is suitable for left-to-right multimodal tasks, it is insufficient for the discrete diffusion models, where medical image and report tokens must be mutually accessible to achieve cross-modal alignment.
As shown in Fig.~\ref{fig_abla}, causal attention leads to blurred boundaries in generated medical images and semantically inconsistent or disorganized content in generated reports.
To address this issue, we remove the causal mask and enable bidirectional attention across the entire sequence to enable fully bidirectional information flow, which is essential for cross-modal consistency. 
% This modification allows medical image tokens and report tokenizations to interact with each other and strengthen cross-modal consistency.

\paragraph{Timestep Embeddings.}
In discrete diffusion, the timestep determines the transition probabilities in the forward process (e.g., the probability of retaining a symbol or replacing it with the [MASK] token). 
The timestep provides the MLLM with essential transition-schedule information in the reverse diffusion. 
% Different timesteps correspond to varying transition probabilities and generation difficulties. 
Without explicitly modeling temporal information, the MLLM backbone cannot recognize the current diffusion stage, which hinders its ability to apply appropriate discrete denoising and alignment strategies during the reverse process. 
Thus, we map each diffusion timestep into a continuous embedding vector and inject it into the MLLM backbone. 
The time embedding further modulates intermediate layers through AdaLN. 
This design ensures that the backbone is aware of the current stage in the reverse transition process.

\paragraph{AdaLN Designs.}
We further incorporate adaptive layer normalization (AdaLN)~\citep{perez2018film} and its variant AdaLN-Zero~\citep{Peebles2022DiT} to enhance the stability and cross-modal consistency of our MLLM backbone.
Unlike standard layer normalization with fixed affine transformations, AdaLN dynamically predicts the normalization parameters from timestep embeddings, ensuring that medical image tokens and report tokenizations are normalized under a shared yet context-aware transformation.
%To further stabilize training, AdaLN-Zero extends this design by introducing a zero-initialized residual scaling parameter $\alpha$, applied before the residual connection, alongside the learned scale and shift parameters $\gamma$ and $\beta$.
%This initialization strategy prevents early-stage instability while still allowing effective conditioning to emerge during optimization.
%Together, AdaLN and AdaLN-Zero provide a computationally efficient normalization mechanism that promotes robust cross-modal alignment throughout the discrete diffusion process.

\section{Experiments}
MeDiM can be applied to varied medical applications: medical image generation, medical report generation, and joint medical image–report pair generation. 
In the following sections, we evaluate MeDiM on these medical tasks and their impact on downstream tasks.
We further discuss the impact of backbone choices in discrete diffusion models and validate the effectiveness of our architectural designs for medical multimodal generation in Sec.~\ref{sec_abla}.

\subsection{Dataset} 
For training and evaluation, we adopt two widely used medical image–text datasets: the MIMIC-CXR~\citep{johnson2019mimic}, a comprehensive chest X-ray with radiology reports, and the PathGen dataset~\citep{sun2024pathgen}, a large-scale collection of pathology image–text pairs. 
Specifically, we use 368,960 chest X-ray pairs from MIMIC-CXR and 736,188 pathology pairs from PathGen, the pathology pairs are subsampled to balance the data distribution. For evaluation, we used 8,000 pathology pairs and adopted the MIMIC-CXR test set.
Both datasets are employed for unified training or supervised fine-tuning (SFT) of MeDiM and baselines.
This unified setting provides a comprehensive evaluation protocol in medical multimodal generation tasks. 

\begin{figure*}[!t]
    \centering
    \setlength{\abovecaptionskip}{1mm} %调整caption与图的距离
    \setlength{\belowcaptionskip}{-5mm}%调整caption与下文的距离
    \includegraphics[width=1\linewidth]{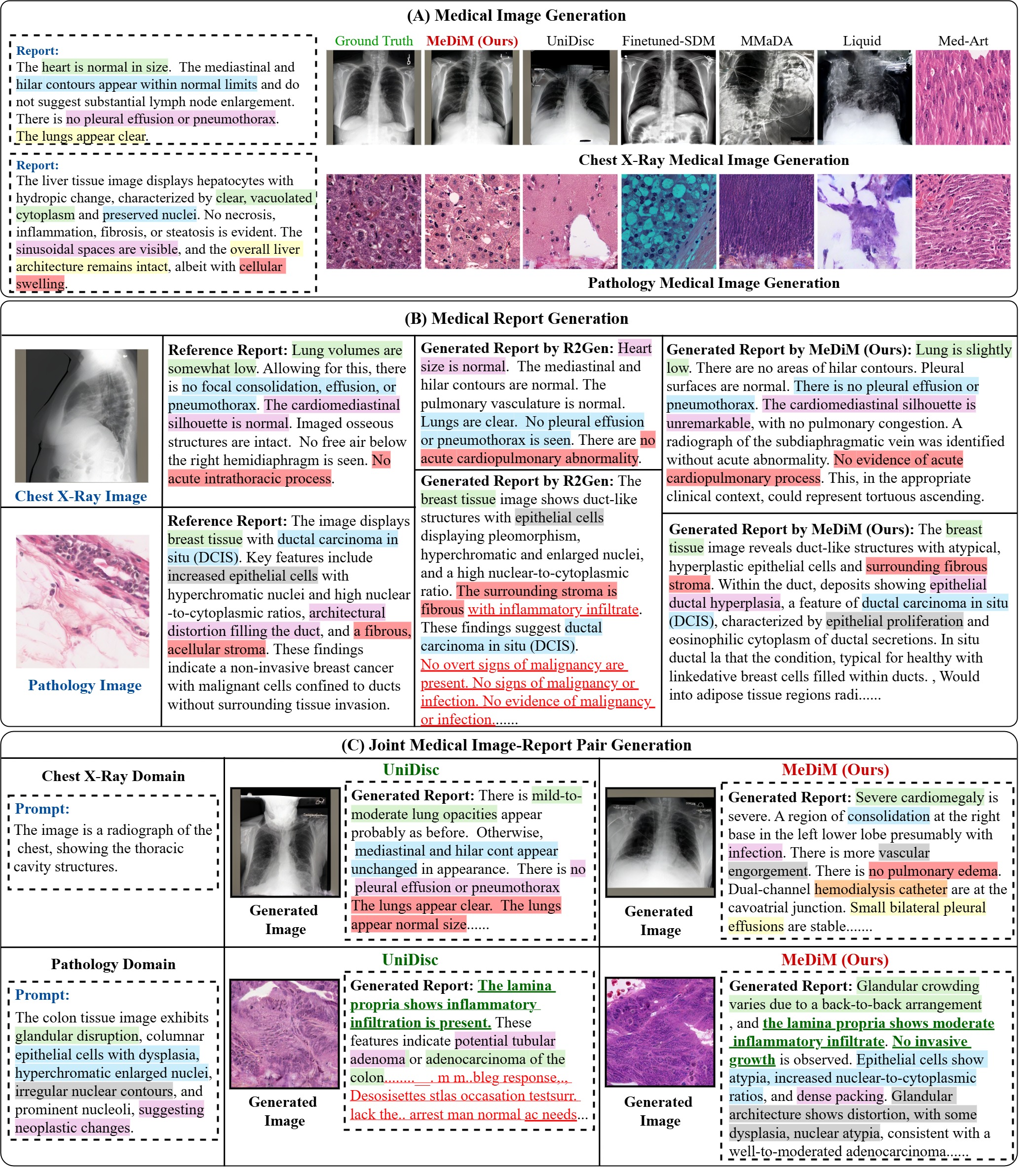}
     \caption{\textbf{Visual comparison of MeDiM against baselines on three tasks:} (A) medical image generation (unique colors indicate the alignment between the reference report and the images generated by MeDiM), (B) medical report generation (generated report and the reference are highlighted with the same colors for matched content, while incorrect content is highlighted with \textcolor{myred}{red underlines}), and (C) joint medical image–report pair generation (generated report and the prompt are highlight with the same colors for matched content, with \textcolor{darkgreen}{green underlines} denoting additional correct content consistent with the image, and \textcolor{myred}{red underlines} marking incorrect content.).}
     \label{fig_qual}
\end{figure*}

\subsection{Settings and Metrics}
We adopt pretrained Liquid~\citep{wu2024liquid} as the MLLM backbone of MeDiM, combined with the VQGAN~\citep{esser2020taming} encoder from Chameleon~\citep{team2024chameleon} for image tokenization and the LLaMA tokenizer~\citep{touvron2023llama} for text processing. The model is trained for 1M steps with a Warmup Cosine Annealing with Restarts learning rate schedule on 8 A100 GPUs, starting from $1\times10^{-5}$. Training uses images at a resolution of $512 \times 512$ and text sequences truncated to a maximum of 256 tokens. During inference, MeDiM follows the MaskGIT inference strategy~\citep{chang2022maskgit}, progressively refining [MASK] tokens to generate coherent multimodal medical output. 
For medical image generation, we evaluate MeDiM using Fréchet inception distance (FID) and inception score (IS).
For medical report generation, we adopt standard natural language generation metrics, including BLEU (B-1, B-2, B-3), METEOR (MTR), and ROUGE-L (R-L).
For joint image–report pair generation, we assess cross-modal consistency using the Qwen2-VL~\citep{qwen2.5} as an automatic evaluator, complemented by human evaluation. And we evaluate downstream task performance to further verify the practical utility of the generated multimodal outputs in medical scenarios.

\begin{table}
\centering
\begin{minipage}[t]{0.48\textwidth}
   \centering
    
\caption{Quantitative comparison for chest X-ray image generation. * denotes fine-tuned models.}
\vspace{-0.1in}
\small
\begin{tabular}{cc cc} % 去掉前两列后的竖线
\toprule
Method & Type & FID $\downarrow$ & IS $\uparrow$ \\
\midrule
SDM & NA & 120.28 & 2.92 \\
SDM (SFT) & NA & 78.97 & 2.91 \\
\midrule
UniDisc & DDM & 82.54 & 2.82 \\
\midrule
U-KAN & MED & 94.58 & 2.89 \\
Med-Art & MED & 168.92 & \textbf{3.82} \\
\midrule
Liquid* & MLLM & 156.09 & 1.97 \\
MMaDA* & MLLM & 134.01 & 2.05\\
\midrule
MeDiM (Ours) & MLLM & \textbf{16.60} & 2.87 \\
\bottomrule
\end{tabular}
    \label{tab_1}
   \end{minipage}
   % \vspace{0.1in}
   \hspace{0.08in}
   \begin{minipage}[t]{0.46\textwidth}
   
\caption{Quantitative comparison for pathology image generation. * denotes fine-tuned models.}
\vspace{-0.1in}
\small
\label{tab_2}
\begin{tabular}{cc cc} % 去掉前两列后的竖线
\toprule
Method & Type & FID $\downarrow$ & IS $\uparrow$ \\
\midrule
SDM & NA & 159.93 & 2.59 \\
SDM (SFT) & NA & 55.76 & 4.03 \\
\midrule
UniDisc & DDM & 80.99 & 4.15 \\
\midrule
U-KAN & MED & 73.76 & 2.98 \\
Med-Art & MED & 107.45 & 2.93 \\
\midrule
Liquid* & MLLM & 171.17 & 3.07 \\
MMaDA* & MLLM & 155.76 & 3.64 \\
\midrule
MeDiM (Ours) & MLLM & \textbf{24.19} & \textbf{4.28} \\
\bottomrule
\end{tabular}
    % \label{tab3}
  \end{minipage}
\vspace{-0.2in}
\end{table}

\subsection{Medical Image Generation}
For the medical image generation task, MeDiM retains all report tokenizations while replacing the image sequence with states. Thus, the generated images are conditioned on the corresponding reports. 
We compare MeDiM against three categories of baselines:
(1) natural diffusion models (NA), such as the stable diffusion model (SDM)~\citep{rombach2022high};
(2) specialized medical image generation models (MED), including U-KAN~\citep{li2025u} and Med-Art~\citep{guo2025med} (Notably, we introduce the class embedding of DiT into U-KAN, which enabled unified medical image generation.); (3) DDM: UniDisc~\citep{swerdlow2025unified}, and (4) multimodal generation–understanding models (MLLM), such as Liquid~\citep{wu2024liquid} and MMaDA~\citep{yang2025mmada}. 

\paragraph{Results.}
Tab.~\ref{tab_1} and Tab.~\ref{tab_2} show the evaluation results of medical report–to–image generation across different medical benchmarks. 
In comparison to baselines, MeDiM demonstrates consistent SoTA performance. 
In Fig.~\ref{fig_qual}, both MMaDA and Liquid, which were fine-tuned on medical image–report pairs, produce implausible distortions in the generated medical images, due to domain and task shift. 
%Med-Art overfitting to pathology image generation. 
The fine-tuned SDM fails to maintain high-fidelity results in medical image generation tasks and shows noticeable color shifts in pathology image synthesis. 
Although UniDisc can be applied to diverse medical image generation, the generated outputs are not always consistent with the corresponding reports (e.g., large blurred shadows in the lower lungs of chest X-rays while the reports describe “\textit{The lungs appear clear}''). 
Compared to baselines, MeDiM generates medical images that exhibit higher fidelity and greater consistency with the medical reports. 

\begin{table*}[ht]
\centering
\caption{Comparison of our method (MeDiM) with different types of baselines on MIMIC-CXR and PathGen datasets. * indicates models that are fine-tuned under our dataset setting.}
\vspace{-0.1in}
\label{tab_report}
\small
\resizebox{\textwidth}{!}{
\begin{tabular}{cc ccccc ccccc} % 去掉前两列后的竖线
\toprule
\multirow{2}{*}{Method} & \multirow{2}{*}{Type} 
& \multicolumn{5}{c}{MIMIC-CXR} 
& \multicolumn{5}{c}{PathGen} \\
\cmidrule(lr){3-7} \cmidrule(lr){8-12}
& & B-1 & B-2 & B-3 & MTR & R-L & B-1 & B-2 & B-3 & MTR & R-L \\
\midrule
BLIP & NA & 0.240 & 0.125 & 0.053 & 0.125 & 0.265 & 0.106 & 0.054 & 0.031 & 0.140 & 0.236 \\
\midrule
R2Gen & MED & 0.305 & 0.179 & 0.104 & 0.233 & \textbf{0.395} & 0.160 & 0.090 & \textbf{0.055} & 0.251 & \textbf{0.278} \\
R2GenCMN & MED & 0.266 & 0.132 & 0.061 & 0.223 & 0.225 & 0.142 & 0.069 & 0.037 & 0.248 & 0.267 \\
BLLM & MED & 0.252 & 0.152 & 0.070 & 0.201 & 0.220 & 0.113 & 0.053 & 0.018 & 0.154 & 0.229 \\
\midrule
UniDisc & DDM & 0.270 & 0.137 & 0.075 & 0.224 & 0.206 & 0.109 & 0.039 & 0.012 & 0.180 & 0.113 \\
\midrule
Liquid* & MLLM & 0.186 & 0.104 & 0.037 & 0.170 & 0.172 & 0.124 & 0.028 & 0.009 & 0.107 & 0.121 \\
MMaDA* & MLLM & 0.153 & 0.102 & 0.031 & 0.164 & 0.185 & 0.172 & \textbf{0.108} & 0.052 & 0.200 & 0.258 \\
\midrule
MeDiM (Ours) & MLLM & \textbf{0.328} & \textbf{0.185} & \textbf{0.109} & \textbf{0.265} & 0.297 & \textbf{0.185} &  0.084 & 0.037 & \textbf{0.258} & 0.226 \\
\bottomrule
\end{tabular}
}
\end{table*}
\vspace{-0.1in}

\subsection{Medical Report Generation}
%In the general medical report generation task, MeDiM generates reports based on medical images by masking the report sequence with absorbing states while preserving image tokens. 
Diverse baselines are compared in the medical report generation task (NA), including BLIP~\citep{li2022blip} as a representative of general-purpose captioning models, R2Gen~\citep{chen-emnlp-2020-r2gen}, R2GenCMN~\citep{chen-acl-2021-r2gencmn}, and BLLM~\citep{liu2024bootstrapping} as specialized medical report generation systems (MED), UniDisc~\citep{swerdlow2025unified} in DDM, and Liquid~\citep{wu2024liquid} and MMaDA~\citep{yang2025mmada} as multimodal generation–understanding models (MLLM). 
To further enhance the reliability of report evaluation in PathGen, we employ Qwen2-VL to filter the test set and select 5,000 high-quality medical reports, which are used as the ground truth (GT). 

\paragraph{Results.}
Tab.~\ref{tab_report} shows the performance of MeDiM on multiple benchmarks for medical image report generation. 
Compared with MLLM-based generative-understanding models such as Liquid, MMaDA, and UniDisc, as well as BLIP, a natural image captioning model, our approach achieves superior performance on both benchmarks. 
In comparison with specialized medical report generation approaches, such as R2Gen, R2GenCMN, and BLLM, our method also attains comparable or superior results. 
To further complement the quantitative evaluations, Fig.~\ref{fig_qual} presents qualitative comparisons of MeDiM and R2Gen. 
The R2Gen suffers from notable deficiencies, including semantic repetition, omission of salient details, and logical inconsistencies (see Fig.~\ref{fig_qual}).
In contrast, MeDiM reduces semantic redundancy and faithfully interprets clinically significant details.

\subsection{Joint Medical Image-Report Pair Generation}
In the joint medical image–report pair generation task, only UniDisc and MeDiM can simultaneously generate medical images and their corresponding reports.
We adopt UniDisc as the baseline and evaluate the paired generation results in cross-modal consistency and downstream task performance. 

\vspace{-0.1in}
\begin{figure*}[htbp]
    \centering
    % 第一张
    \begin{subfigure}[t]{0.32\textwidth}
        \centering
        \includegraphics[width=\linewidth]{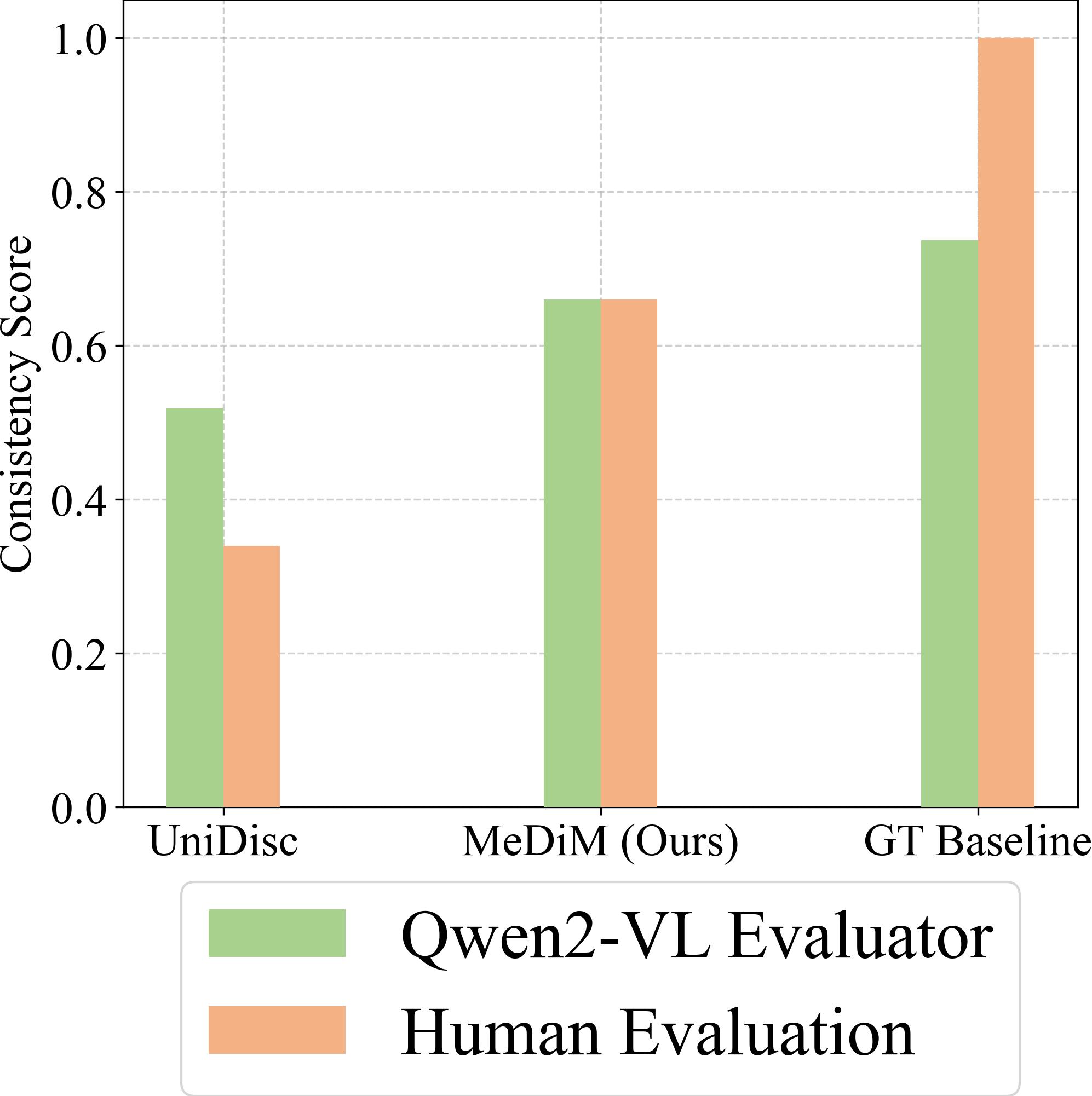}
        \caption{Image–Report alignment evaluation.}
        \label{fig_joint_a}
    \end{subfigure}
    \hfill
    % 第二张
    \begin{subfigure}[t]{0.32\textwidth}
        \centering
        \includegraphics[width=\linewidth]{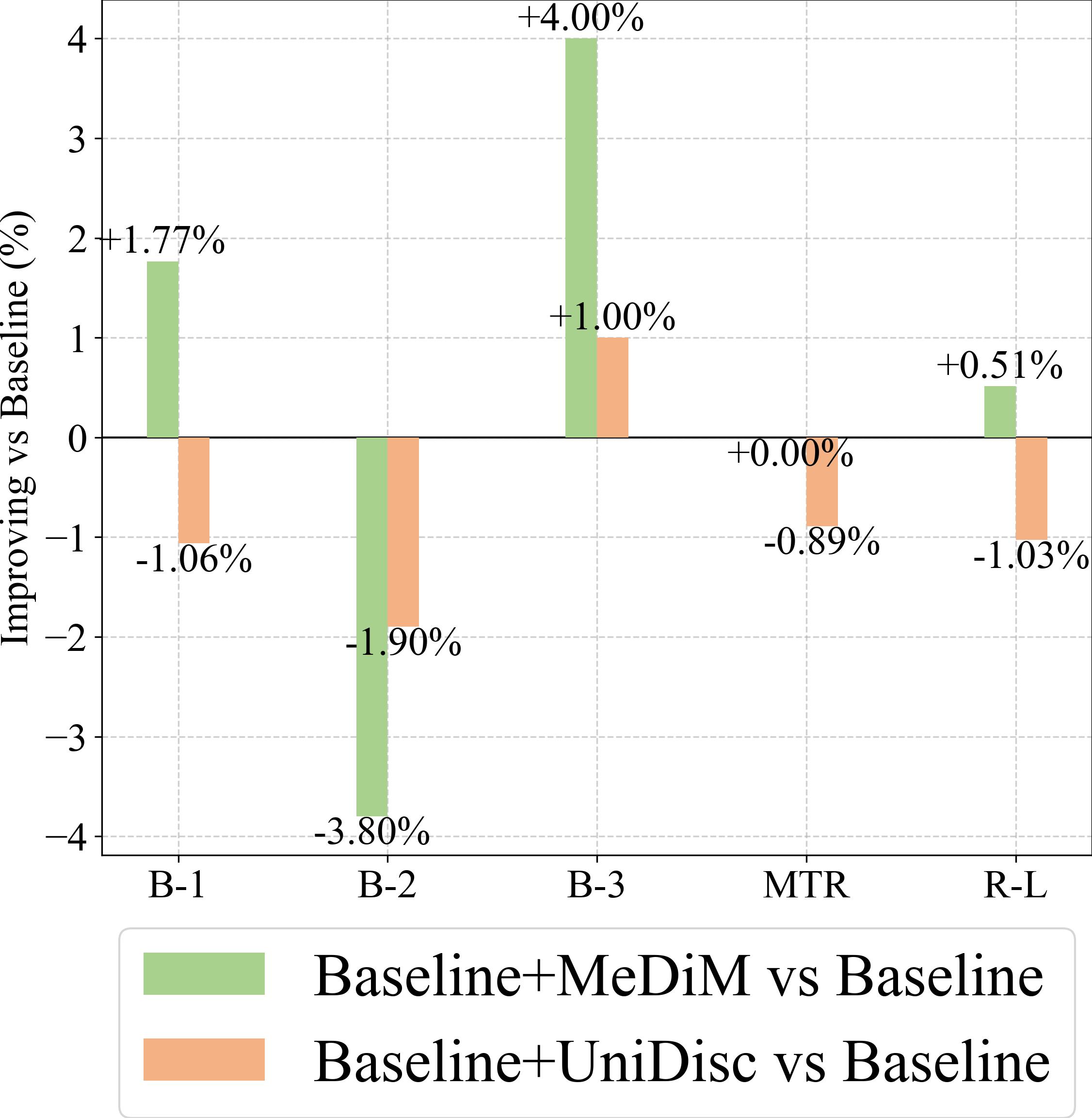}
        \caption{Downstream task evaluation in MIMIC-CXR}
        \label{fig_joint_b}
    \end{subfigure}
    \hfill
    % 第三张
    \begin{subfigure}[t]{0.32\textwidth}
        \centering
        \includegraphics[width=\linewidth]{{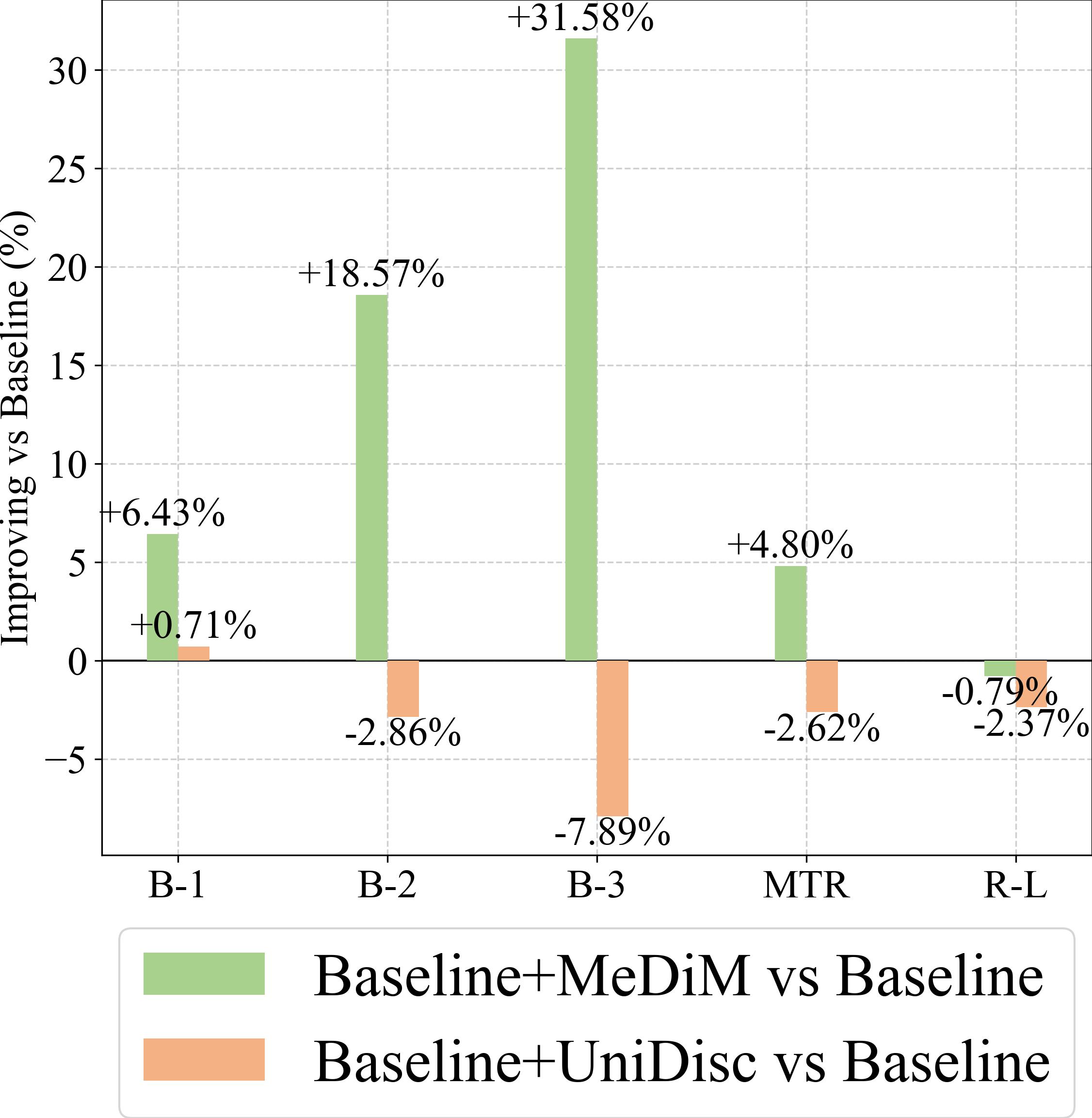}}
        \caption{Downstream task evaluation in PathGen}
        \label{fig_joint_c}
    \end{subfigure}
    % \vspace{-0.1in}
    \caption{Quantitative evaluation of MeDiM on the joint medical image–report generation task.}
    \label{fig_joint}
\end{figure*}
\vspace{-0.2in}

\paragraph{Alignment.}
We evaluate the consistency of generated medical image–report pairs using the large VLM Qwen2-VL. 
We employ [MASK] sequences as inputs, unconditionally generate 8,000 pathology image–report pairs and 5,159 chest X-ray image–report pairs to support consistency evaluation.
To further ensure reliability, we complement this with human evaluation, conducted on 100 unconditionally generated image–report pairs sampled in a 1:1 ratio between pathology and chest X-ray domains. 
As illustrated in Fig.~\ref{fig_joint_a}, large VLMs and human evaluators produce consistent judgments, both showing that MeDiM attains high confidence. 
Meanwhile, as observed in Fig.~\ref{fig_qual}, MeDiM generates image–report pairs that adhere to the prompts and remain semantically consistent. 
%Beyond the prompts, MeDiM can generate new medical descriptions. For example, in pathology image–report synthesis, MeDiM produced the description independent of prompt “\textit{the lamina propria shows moderate inflammatory infiltrate. No invasive growth}”, highlighting its practical value. 
%While UniDisc demonstrates consistency in chest X-ray image–report synthesis, it tends to generate semantically disordered content in pathology report generation, particularly as the length of the generated reports increases. 
%These findings indicate that MeDiM has the potential to generate clinically reliable and semantically consistent medical image–report pairs.

\paragraph{Downstream Task.}
In the downstream evaluation, we investigate the impact of synthetic medical image–report pairs on VLM performance under low-data medical settings.
We first construct simple text prompts based on image modality, anatomical region, and pathology condition. These prompts are injected into the [MASK]-initialized sequences during inference, thereby guiding MeDiM or UniDisc in conditionally generating 200k medical image–report pairs with a 1:1 balanced distribution.
Then, we sample 200k real pairs from the MIMIC-CXR and PathGen training sets in a 1:1 ratio and merge them with the 200k synthetic pairs to form a balanced dataset.
This dataset is used to train the medical report generation baseline R2Gen. 
As shown in Fig.~\ref{fig_joint_b} and Fig.~\ref{fig_joint_c}, the image–report pairs generated by MeDiM lead to noticeable gains in pathology visual analysis tasks.

\section{Conclusion}

In this work, we introduce \textbf{MeDiM}, the first medical discrete diffusion model designed to address the critical need for a unified framework that can generate multimodal medical data across different domains. By learning shared probabilistic distributions, MeDiM acts as a flexible multimodal generator without requiring modality-specific components. We adapt a MLLM as the diffusion backbone, leveraging its powerful distributional alignment priors to create a unified representation space. Extensive experiments demonstrate that MeDiM achieves robust, state-of-the-art performance in medical image generation, report generation, and joint image-text synthesis. Crucially, MeDiM pioneers a new class of paired image-report synthesis, creating consistent data that improves the performance of downstream vision-language models. This highlights the effectiveness of MLLM as the multimodal diffusion backbone, which can be a promising step towards building versatile, generalist medical AI agents.
%However, MeDiM still lags behind specialized expert medical models, and future work will focus on integrating medical knowledge–aware MLLM backbones, thus extending its applicability to broader clinical settings.

\section{Acknowledgement}
This work was partially funded by an unrestricted gift from Google.

\bibliography{iclr2026_conference}

\begin{thebibliography}{67}
\providecommand{\natexlab}[1]{#1}
\providecommand{\url}[1]{\texttt{#1}}
\expandafter\ifx\csname urlstyle\endcsname\relax
  \providecommand{\doi}[1]{doi: #1}\else
  \providecommand{\doi}{doi: \begingroup \urlstyle{rm}\Url}\fi

\bibitem[Austin et~al.(2021)Austin, Johnson, Ho, Tarlow, and Van Den~Berg]{austin2021structured}
Jacob Austin, Daniel~D Johnson, Jonathan Ho, Daniel Tarlow, and Rianne Van Den~Berg.
\newblock Structured denoising diffusion models in discrete state-spaces.
\newblock \emph{Advances in neural information processing systems}, 34:\penalty0 17981--17993, 2021.

\bibitem[Brock et~al.(2018)Brock, Donahue, and Simonyan]{brock2018large}
Andrew Brock, Jeff Donahue, and Karen Simonyan.
\newblock Large scale gan training for high fidelity natural image synthesis.
\newblock \emph{arXiv preprint arXiv:1809.11096}, 2018.

\bibitem[Chambon et~al.(2022)Chambon, Bluethgen, Delbrouck, Van~der Sluijs, Po{\l}acin, Chaves, Abraham, Purohit, Langlotz, and Chaudhari]{chambon2022roentgen}
Pierre Chambon, Christian Bluethgen, Jean-Benoit Delbrouck, Rogier Van~der Sluijs, Ma{\l}gorzata Po{\l}acin, Juan Manuel~Zambrano Chaves, Tanishq~Mathew Abraham, Shivanshu Purohit, Curtis~P Langlotz, and Akshay Chaudhari.
\newblock Roentgen: vision-language foundation model for chest x-ray generation.
\newblock \emph{arXiv preprint arXiv:2211.12737}, 2022.

\bibitem[Chang et~al.(2022)Chang, Zhang, Jiang, Liu, and Freeman]{chang2022maskgit}
Huiwen Chang, Han Zhang, Lu~Jiang, Ce~Liu, and William~T Freeman.
\newblock Maskgit: Masked generative image transformer.
\newblock In \emph{Proceedings of the IEEE/CVF conference on computer vision and pattern recognition}, pp.\  11315--11325, 2022.

\bibitem[Chartsias et~al.(2017)Chartsias, Joyce, Dharmakumar, and Tsaftaris]{chartsias2017adversarial}
Agisilaos Chartsias, Thomas Joyce, Rohan Dharmakumar, and Sotirios~A Tsaftaris.
\newblock Adversarial image synthesis for unpaired multi-modal cardiac data.
\newblock In \emph{International workshop on simulation and synthesis in medical imaging}, pp.\  3--13. Springer, 2017.

\bibitem[Chen et~al.(2023)Chen, Shen, Lin, Luo, Li, and Yuan]{chen2023fine}
Wenting Chen, Linlin Shen, Jingyang Lin, Jiebo Luo, Xiang Li, and Yixuan Yuan.
\newblock Fine-grained image-text alignment in medical imaging enables explainable cyclic image-report generation.
\newblock \emph{arXiv preprint arXiv:2312.08078}, 2023.

\bibitem[Chen et~al.(2020)Chen, Song, Chang, and Wan]{chen-emnlp-2020-r2gen}
Zhihong Chen, Yan Song, Tsung-Hui Chang, and Xiang Wan.
\newblock Generating radiology reports via memory-driven transformer.
\newblock In \emph{Proceedings of the 2020 Conference on Empirical Methods in Natural Language Processing}, November 2020.

\bibitem[Chen et~al.(2021)Chen, Shen, Song, and Wan]{chen-acl-2021-r2gencmn}
Zhihong Chen, Yaling Shen, Yan Song, and Xiang Wan.
\newblock Generating radiology reports via memory-driven transformer.
\newblock In \emph{Proceedings of the Joint Conference of the 59th Annual Meeting of the Association for Computational Linguistics and the 11th International Joint Conference on Natural Language Processing}, August 2021.

\bibitem[Esser et~al.(2020)Esser, Rombach, and Ommer]{esser2020taming}
Patrick Esser, Robin Rombach, and Björn Ommer.
\newblock Taming transformers for high-resolution image synthesis, 2020.

\bibitem[Ge et~al.(2023)Ge, Ge, Zeng, Wang, and Shan]{ge2307planting}
Y~Ge, Y~Ge, Z~Zeng, X~Wang, and Y~Shan.
\newblock Planting a seed of vision in large language model. arxiv 2023.
\newblock \emph{arXiv preprint arXiv:2307.08041}, 2023.

\bibitem[Ge et~al.(2024)Ge, Zhao, Zhu, Ge, Yi, Song, Li, Ding, and Shan]{ge2024seed}
Yuying Ge, Sijie Zhao, Jinguo Zhu, Yixiao Ge, Kun Yi, Lin Song, Chen Li, Xiaohan Ding, and Ying Shan.
\newblock Seed-x: Multimodal models with unified multi-granularity comprehension and generation.
\newblock \emph{arXiv preprint arXiv:2404.14396}, 2024.

\bibitem[G{\"u}ng{\"o}r et~al.(2023)G{\"u}ng{\"o}r, Dar, {\"O}zt{\"u}rk, Korkmaz, Bedel, Elmas, Ozbey, and {\c{C}}ukur]{gungor2023adaptive}
Alper G{\"u}ng{\"o}r, Salman~UH Dar, {\c{S}}aban {\"O}zt{\"u}rk, Yilmaz Korkmaz, Hasan~A Bedel, Gokberk Elmas, Muzaffer Ozbey, and Tolga {\c{C}}ukur.
\newblock Adaptive diffusion priors for accelerated mri reconstruction.
\newblock \emph{Medical image analysis}, 88:\penalty0 102872, 2023.

\bibitem[Guo et~al.(2025{\natexlab{a}})Guo, Christensen, and Hannemose]{guo2025med}
Changlu Guo, Anders~Nymark Christensen, and Morten~Rieger Hannemose.
\newblock Med-art: Diffusion transformer for 2d medical text-to-image generation.
\newblock \emph{arXiv preprint arXiv:2506.20449}, 2025{\natexlab{a}}.

\bibitem[Guo et~al.(2020)Guo, Wang, Yasarla, Zhou, Patel, and Jiang]{guo2020anatomic}
Pengfei Guo, Puyang Wang, Rajeev Yasarla, Jinyuan Zhou, Vishal~M Patel, and Shanshan Jiang.
\newblock Anatomic and molecular mr image synthesis using confidence guided cnns.
\newblock \emph{IEEE transactions on medical imaging}, 40\penalty0 (10):\penalty0 2832--2844, 2020.

\bibitem[Guo et~al.(2025{\natexlab{b}})Guo, Zhao, Yang, Xu, Nath, Tang, Simon, Belue, Harmon, Turkbey, et~al.]{guo2025maisi}
Pengfei Guo, Can Zhao, Dong Yang, Ziyue Xu, Vishwesh Nath, Yucheng Tang, Benjamin Simon, Mason Belue, Stephanie Harmon, Baris Turkbey, et~al.
\newblock Maisi: Medical ai for synthetic imaging.
\newblock In \emph{2025 IEEE/CVF Winter Conference on Applications of Computer Vision (WACV)}, pp.\  4430--4441. IEEE, 2025{\natexlab{b}}.

\bibitem[Ho et~al.(2020)Ho, Jain, and Abbeel]{ho2020denoising}
Jonathan Ho, Ajay Jain, and Pieter Abbeel.
\newblock Denoising diffusion probabilistic models.
\newblock \emph{Advances in neural information processing systems}, 33:\penalty0 6840--6851, 2020.

\bibitem[Hu et~al.(2022)Hu, Zheng, Zheng, Cham, Wang, Yang, Tao, and Suganthan]{hu2022unified}
Minghui Hu, Chuanxia Zheng, Heliang Zheng, Tat-Jen Cham, Chaoyue Wang, Zuopeng Yang, Dacheng Tao, and Ponnuthurai~N Suganthan.
\newblock Unified discrete diffusion for simultaneous vision-language generation.
\newblock \emph{arXiv}, 2022.

\bibitem[Huo et~al.(2018)Huo, Xu, Moon, Bao, Assad, Moyo, Savona, Abramson, and Landman]{huo2018synseg}
Yuankai Huo, Zhoubing Xu, Hyeonsoo Moon, Shunxing Bao, Albert Assad, Tamara~K Moyo, Michael~R Savona, Richard~G Abramson, and Bennett~A Landman.
\newblock Synseg-net: Synthetic segmentation without target modality ground truth.
\newblock \emph{IEEE transactions on medical imaging}, 38\penalty0 (4):\penalty0 1016--1025, 2018.

\bibitem[Jiang et~al.(2025)Jiang, Imran, Zhang, Zhou, Liang, Gong, and Shao]{jiang2025fast}
Hongxu Jiang, Muhammad Imran, Teng Zhang, Yuyin Zhou, Muxuan Liang, Kuang Gong, and Wei Shao.
\newblock Fast-ddpm: Fast denoising diffusion probabilistic models for medical image-to-image generation.
\newblock \emph{IEEE Journal of Biomedical and Health Informatics}, 2025.

\bibitem[Johnson et~al.(2019)Johnson, Pollard, Greenbaum, Lungren, Deng, Peng, Lu, Mark, Berkowitz, and Horng]{johnson2019mimic}
Alistair~EW Johnson, Tom~J Pollard, Nathaniel~R Greenbaum, Matthew~P Lungren, Chih-ying Deng, Yifan Peng, Zhiyong Lu, Roger~G Mark, Seth~J Berkowitz, and Steven Horng.
\newblock Mimic-cxr-jpg, a large publicly available database of labeled chest radiographs.
\newblock \emph{arXiv preprint arXiv:1901.07042}, 2019.

\bibitem[Karras et~al.(2019)Karras, Laine, and Aila]{karras2019style}
Tero Karras, Samuli Laine, and Timo Aila.
\newblock A style-based generator architecture for generative adversarial networks.
\newblock In \emph{Proceedings of the IEEE/CVF conference on computer vision and pattern recognition}, pp.\  4401--4410, 2019.

\bibitem[Li et~al.(2025)Li, Liu, Li, Wang, Liu, Liu, Chen, and Yuan]{li2025u}
Chenxin Li, Xinyu Liu, Wuyang Li, Cheng Wang, Hengyu Liu, Yifan Liu, Zhen Chen, and Yixuan Yuan.
\newblock U-kan makes strong backbone for medical image segmentation and generation.
\newblock In \emph{Proceedings of the AAAI Conference on Artificial Intelligence}, volume~39, pp.\  4652--4660, 2025.

\bibitem[Li et~al.(2022)Li, Li, Xiong, and Hoi]{li2022blip}
Junnan Li, Dongxu Li, Caiming Xiong, and Steven Hoi.
\newblock Blip: Bootstrapping language-image pre-training for unified vision-language understanding and generation.
\newblock In \emph{ICML}, 2022.

\bibitem[Li et~al.(2023)Li, Li, Savarese, and Hoi]{li2023blip}
Junnan Li, Dongxu Li, Silvio Savarese, and Steven Hoi.
\newblock Blip-2: Bootstrapping language-image pre-training with frozen image encoders and large language models.
\newblock In \emph{International conference on machine learning}, pp.\  19730--19742. PMLR, 2023.

\bibitem[Liu et~al.(2024{\natexlab{a}})Liu, Tian, Chen, Song, and Zhang]{liu2024bootstrapping}
Chang Liu, Yuanhe Tian, Weidong Chen, Yan Song, and Yongdong Zhang.
\newblock Bootstrapping large language models for radiology report generation.
\newblock In \emph{Proceedings of the AAAI Conference on Artificial Intelligence}, volume~38, pp.\  18635--18643, 2024{\natexlab{a}}.

\bibitem[Liu et~al.(2024{\natexlab{b}})Liu, Yan, Zaharia, and Abbeel]{liu2024world}
Hao Liu, Wilson Yan, Matei Zaharia, and Pieter Abbeel.
\newblock World model on million-length video and language with ringattention. arxiv e-prints, pages arxiv--2402.
\newblock \emph{arXiv preprint arXiv:2402.08268}, 2024{\natexlab{b}}.

\bibitem[Liu et~al.(2024{\natexlab{c}})Liu, Ma, Liu, Jiao, Kang, Miao, and Xie]{liu2024factual}
Kang Liu, Zhuoqi Ma, Mengmeng Liu, Zhicheng Jiao, Xiaolu Kang, Qiguang Miao, and Kun Xie.
\newblock Factual serialization enhancement: A key innovation for chest x-ray report generation.
\newblock \emph{arXiv preprint arXiv:2405.09586}, 2024{\natexlab{c}}.

\bibitem[Liu et~al.(2025)Liu, Ma, Kang, Li, Xie, Jiao, and Miao]{liu2025enhanced}
Kang Liu, Zhuoqi Ma, Xiaolu Kang, Yunan Li, Kun Xie, Zhicheng Jiao, and Qiguang Miao.
\newblock Enhanced contrastive learning with multi-view longitudinal data for chest x-ray report generation.
\newblock In \emph{Proceedings of the Computer Vision and Pattern Recognition Conference}, pp.\  10348--10359, 2025.

\bibitem[Liu et~al.(2024{\natexlab{d}})Liu, Wang, Vaidya, Ruehle, Halverson, Solja{\v{c}}i{\'c}, Hou, and Tegmark]{liu2024kan}
Ziming Liu, Yixuan Wang, Sachin Vaidya, Fabian Ruehle, James Halverson, Marin Solja{\v{c}}i{\'c}, Thomas~Y Hou, and Max Tegmark.
\newblock Kan: Kolmogorov-arnold networks.
\newblock \emph{arXiv preprint arXiv:2404.19756}, 2024{\natexlab{d}}.

\bibitem[Lou et~al.(2023)Lou, Meng, and Ermon]{lou2023discrete}
Aaron Lou, Chenlin Meng, and Stefano Ermon.
\newblock Discrete diffusion modeling by estimating the ratios of the data distribution.
\newblock \emph{arXiv preprint arXiv:2310.16834}, 2023.

\bibitem[Lu et~al.(2024)Lu, Clark, Lee, Zhang, Khosla, Marten, Hoiem, and Kembhavi]{lu2024unified}
Jiasen Lu, Christopher Clark, Sangho Lee, Zichen Zhang, Savya Khosla, Ryan Marten, Derek Hoiem, and Aniruddha Kembhavi.
\newblock Unified-io 2: Scaling autoregressive multimodal models with vision language audio and action.
\newblock In \emph{Proceedings of the IEEE/CVF Conference on Computer Vision and Pattern Recognition}, pp.\  26439--26455, 2024.

\bibitem[Lyu \& Wang(2022)Lyu and Wang]{lyu2022conversion}
Qing Lyu and Ge~Wang.
\newblock Conversion between ct and mri images using diffusion and score-matching models.
\newblock \emph{arXiv preprint arXiv:2209.12104}, 2022.

\bibitem[Mao et~al.(2025)Mao, Wang, Tang, Xu, Wang, Yang, Zhou, and Zhou]{mao2025medsegfactory}
Jiawei Mao, Yuhan Wang, Yucheng Tang, Daguang Xu, Kang Wang, Yang Yang, Zongwei Zhou, and Yuyin Zhou.
\newblock Medsegfactory: Text-guided generation of medical image-mask pairs.
\newblock \emph{arXiv preprint arXiv:2504.06897}, 2025.

\bibitem[Moor et~al.(2023)Moor, Banerjee, Abad, Krumholz, Leskovec, Topol, and Rajpurkar]{moor2023foundation}
Michael Moor, Oishi Banerjee, Zahra Shakeri~Hossein Abad, Harlan~M Krumholz, Jure Leskovec, Eric~J Topol, and Pranav Rajpurkar.
\newblock Foundation models for generalist medical artificial intelligence.
\newblock \emph{Nature}, 616\penalty0 (7956):\penalty0 259--265, 2023.

\bibitem[Nie et~al.(2025)Nie, Zhu, You, Zhang, Ou, Hu, Zhou, Lin, Wen, and Li]{nie2025large}
Shen Nie, Fengqi Zhu, Zebin You, Xiaolu Zhang, Jingyang Ou, Jun Hu, Jun Zhou, Yankai Lin, Ji-Rong Wen, and Chongxuan Li.
\newblock Large language diffusion models.
\newblock \emph{arXiv preprint arXiv:2502.09992}, 2025.

\bibitem[Peebles \& Xie(2022)Peebles and Xie]{Peebles2022DiT}
William Peebles and Saining Xie.
\newblock Scalable diffusion models with transformers.
\newblock \emph{arXiv preprint arXiv:2212.09748}, 2022.

\bibitem[Perez et~al.(2018)Perez, Strub, De~Vries, Dumoulin, and Courville]{perez2018film}
Ethan Perez, Florian Strub, Harm De~Vries, Vincent Dumoulin, and Aaron Courville.
\newblock Film: Visual reasoning with a general conditioning layer.
\newblock In \emph{Proceedings of the AAAI conference on artificial intelligence}, volume~32, 2018.

\bibitem[Polamreddy et~al.(2025)Polamreddy, Roy, Yueh, Mahato, Kuppili, Li, and Zhang]{polamreddy2025leapfrog}
Lakshmikar~R Polamreddy, Kalyan Roy, Sheng-Han Yueh, Deepshikha Mahato, Shilpa Kuppili, Jialu Li, and Youshan Zhang.
\newblock Leapfrog latent consistency model (llcm) for medical images generation.
\newblock In \emph{2025 IEEE/ACM Conference on Connected Health: Applications, Systems and Engineering Technologies (CHASE)}, pp.\  430--435. IEEE, 2025.

\bibitem[Radford et~al.(2021)Radford, Kim, Hallacy, Ramesh, Goh, Agarwal, Sastry, Askell, Mishkin, Clark, et~al.]{radford2021learning}
Alec Radford, Jong~Wook Kim, Chris Hallacy, Aditya Ramesh, Gabriel Goh, Sandhini Agarwal, Girish Sastry, Amanda Askell, Pamela Mishkin, Jack Clark, et~al.
\newblock Learning transferable visual models from natural language supervision.
\newblock In \emph{International conference on machine learning}, pp.\  8748--8763. PmLR, 2021.

\bibitem[Rombach et~al.(2022)Rombach, Blattmann, Lorenz, Esser, and Ommer]{rombach2022high}
Robin Rombach, Andreas Blattmann, Dominik Lorenz, Patrick Esser, and Bj{\"o}rn Ommer.
\newblock High-resolution image synthesis with latent diffusion models.
\newblock In \emph{Proceedings of the IEEE/CVF conference on computer vision and pattern recognition}, pp.\  10684--10695, 2022.

\bibitem[Sahoo et~al.(2024)Sahoo, Arriola, Schiff, Gokaslan, Marroquin, Chiu, Rush, and Kuleshov]{sahoo2024simple}
Subham Sahoo, Marianne Arriola, Yair Schiff, Aaron Gokaslan, Edgar Marroquin, Justin Chiu, Alexander Rush, and Volodymyr Kuleshov.
\newblock Simple and effective masked diffusion language models.
\newblock \emph{Advances in Neural Information Processing Systems}, 37:\penalty0 130136--130184, 2024.

\bibitem[Sohl-Dickstein et~al.(2015)Sohl-Dickstein, Weiss, Maheswaranathan, and Ganguli]{sohl2015deep}
Jascha Sohl-Dickstein, Eric Weiss, Niru Maheswaranathan, and Surya Ganguli.
\newblock Deep unsupervised learning using nonequilibrium thermodynamics.
\newblock In \emph{International conference on machine learning}, pp.\  2256--2265. pmlr, 2015.

\bibitem[Sun et~al.(2023)Sun, Yu, Cui, Zhang, Zhang, Wang, Gao, Liu, Huang, and Wang]{sun2023emu}
Quan Sun, Qiying Yu, Yufeng Cui, Fan Zhang, Xiaosong Zhang, Yueze Wang, Hongcheng Gao, Jingjing Liu, Tiejun Huang, and Xinlong Wang.
\newblock Emu: Generative pretraining in multimodality.
\newblock \emph{arXiv preprint arXiv:2307.05222}, 2023.

\bibitem[Sun et~al.(2024{\natexlab{a}})Sun, Cui, Zhang, Zhang, Yu, Wang, Rao, Liu, Huang, and Wang]{sun2024generative}
Quan Sun, Yufeng Cui, Xiaosong Zhang, Fan Zhang, Qiying Yu, Yueze Wang, Yongming Rao, Jingjing Liu, Tiejun Huang, and Xinlong Wang.
\newblock Generative multimodal models are in-context learners.
\newblock In \emph{Proceedings of the IEEE/CVF Conference on Computer Vision and Pattern Recognition}, pp.\  14398--14409, 2024{\natexlab{a}}.

\bibitem[Sun et~al.(2024{\natexlab{b}})Sun, Zhang, Si, Zhu, Shui, Zhang, Li, Lyu, Lin, and Yang]{sun2024pathgen}
Yuxuan Sun, Yunlong Zhang, Yixuan Si, Chenglu Zhu, Zhongyi Shui, Kai Zhang, Jingxiong Li, Xingheng Lyu, Tao Lin, and Lin Yang.
\newblock Pathgen-1.6 m: 1.6 million pathology image-text pairs generation through multi-agent collaboration.
\newblock \emph{arXiv preprint arXiv:2407.00203}, 2024{\natexlab{b}}.

\bibitem[Swerdlow et~al.(2025)Swerdlow, Prabhudesai, Gandhi, Pathak, and Fragkiadaki]{swerdlow2025unified}
Alexander Swerdlow, Mihir Prabhudesai, Siddharth Gandhi, Deepak Pathak, and Katerina Fragkiadaki.
\newblock Unified multimodal discrete diffusion.
\newblock \emph{arXiv preprint arXiv:2503.20853}, 2025.

\bibitem[Tanida et~al.(2023)Tanida, M{\"u}ller, Kaissis, and Rueckert]{tanida2023interactive}
Tim Tanida, Philip M{\"u}ller, Georgios Kaissis, and Daniel Rueckert.
\newblock Interactive and explainable region-guided radiology report generation.
\newblock In \emph{Proceedings of the IEEE/CVF Conference on Computer Vision and Pattern Recognition}, pp.\  7433--7442, 2023.

\bibitem[Team(2024{\natexlab{a}})]{team2024chameleon}
Chameleon Team.
\newblock Chameleon: Mixed-modal early-fusion foundation models.
\newblock \emph{arXiv preprint arXiv:2405.09818}, 2024{\natexlab{a}}.

\bibitem[Team(2024{\natexlab{b}})]{qwen2.5}
Qwen Team.
\newblock Qwen2.5: A party of foundation models, September 2024{\natexlab{b}}.
\newblock URL \url{https://qwenlm.github.io/blog/qwen2.5/}.

\bibitem[Touvron et~al.(2023)Touvron, Lavril, Izacard, Martinet, Lachaux, Lacroix, Rozi{\`e}re, Goyal, Hambro, Azhar, et~al.]{touvron2023llama}
Hugo Touvron, Thibaut Lavril, Gautier Izacard, Xavier Martinet, Marie-Anne Lachaux, Timoth{\'e}e Lacroix, Baptiste Rozi{\`e}re, Naman Goyal, Eric Hambro, Faisal Azhar, et~al.
\newblock Llama: Open and efficient foundation language models.
\newblock \emph{arXiv preprint arXiv:2302.13971}, 2023.

\bibitem[Van Den~Oord et~al.(2017)Van Den~Oord, Vinyals, et~al.]{van2017neural}
Aaron Van Den~Oord, Oriol Vinyals, et~al.
\newblock Neural discrete representation learning.
\newblock \emph{Advances in neural information processing systems}, 30, 2017.

\bibitem[Vaswani et~al.(2017)Vaswani, Shazeer, Parmar, Uszkoreit, Jones, Gomez, Kaiser, and Polosukhin]{vaswani2017attention}
Ashish Vaswani, Noam Shazeer, Niki Parmar, Jakob Uszkoreit, Llion Jones, Aidan~N Gomez, {\L}ukasz Kaiser, and Illia Polosukhin.
\newblock Attention is all you need.
\newblock In \emph{Advances in Neural Information Processing Systems}, pp.\  5998--6008, 2017.

\bibitem[Wu et~al.(2025)Wu, Chen, Wu, Ma, Liu, Pan, Liu, Xie, Yu, Ruan, et~al.]{wu2025janus}
Chengyue Wu, Xiaokang Chen, Zhiyu Wu, Yiyang Ma, Xingchao Liu, Zizheng Pan, Wen Liu, Zhenda Xie, Xingkai Yu, Chong Ruan, et~al.
\newblock Janus: Decoupling visual encoding for unified multimodal understanding and generation.
\newblock In \emph{Proceedings of the Computer Vision and Pattern Recognition Conference}, pp.\  12966--12977, 2025.

\bibitem[Wu et~al.(2024{\natexlab{a}})Wu, Jiang, Ma, Liu, Zhao, Yuan, Bai, and Bai]{wu2024liquid}
Junfeng Wu, Yi~Jiang, Chuofan Ma, Yuliang Liu, Hengshuang Zhao, Zehuan Yuan, Song Bai, and Xiang Bai.
\newblock Liquid: Language models are scalable and unified multi-modal generators.
\newblock \emph{arXiv preprint arXiv:2412.04332}, 2024{\natexlab{a}}.

\bibitem[Wu et~al.(2024{\natexlab{b}})Wu, Zhang, Chen, Tang, Li, Fang, Zhu, Xie, Yin, Yi, et~al.]{wu2024vila}
Yecheng Wu, Zhuoyang Zhang, Junyu Chen, Haotian Tang, Dacheng Li, Yunhao Fang, Ligeng Zhu, Enze Xie, Hongxu Yin, Li~Yi, et~al.
\newblock Vila-u: a unified foundation model integrating visual understanding and generation.
\newblock \emph{arXiv preprint arXiv:2409.04429}, 2024{\natexlab{b}}.

\bibitem[Xie et~al.(2024{\natexlab{a}})Xie, Mao, Bai, Zhang, Wang, Lin, Gu, Chen, Yang, and Shou]{xie2024show}
Jinheng Xie, Weijia Mao, Zechen Bai, David~Junhao Zhang, Weihao Wang, Kevin~Qinghong Lin, Yuchao Gu, Zhijie Chen, Zhenheng Yang, and Mike~Zheng Shou.
\newblock Show-o: One single transformer to unify multimodal understanding and generation.
\newblock \emph{arXiv preprint arXiv:2408.12528}, 2024{\natexlab{a}}.

\bibitem[Xie et~al.(2024{\natexlab{b}})Xie, Chen, Wang, To, Lee, Khoo, Hendy, Koh, Xia, and Wu]{xie2024pairaug}
Yutong Xie, Qi~Chen, Sinuo Wang, Minh-Son To, Iris Lee, Ee~Win Khoo, Kerolos Hendy, Daniel Koh, Yong Xia, and Qi~Wu.
\newblock Pairaug: What can augmented image-text pairs do for radiology?
\newblock In \emph{Proceedings of the IEEE/CVF Conference on Computer Vision and Pattern Recognition}, pp.\  11652--11661, 2024{\natexlab{b}}.

\bibitem[Yang et~al.(2025)Yang, Tian, Li, Zhang, Shen, Tong, and Wang]{yang2025mmada}
Ling Yang, Ye~Tian, Bowen Li, Xinchen Zhang, Ke~Shen, Yunhai Tong, and Mengdi Wang.
\newblock Mmada: Multimodal large diffusion language models.
\newblock \emph{arXiv preprint arXiv:2505.15809}, 2025.

\bibitem[Yang et~al.(2022)Yang, Wu, Ge, Zhou, and Xiao]{yang2022knowledge}
Shuxin Yang, Xian Wu, Shen Ge, S~Kevin Zhou, and Li~Xiao.
\newblock Knowledge matters: Chest radiology report generation with general and specific knowledge.
\newblock \emph{Medical image analysis}, 80:\penalty0 102510, 2022.

\bibitem[Yu et~al.(2023)Yu, Shi, Pasunuru, Muller, Golovneva, Wang, Babu, Tang, Karrer, Sheynin, et~al.]{yu2023scaling}
Lili Yu, Bowen Shi, Ramakanth Pasunuru, Benjamin Muller, Olga Golovneva, Tianlu Wang, Arun Babu, Binh Tang, Brian Karrer, Shelly Sheynin, et~al.
\newblock Scaling autoregressive multi-modal models: Pretraining and instruction tuning.
\newblock \emph{arXiv preprint arXiv:2309.02591}, 2023.

\bibitem[Zeng et~al.(2023)Zeng, Zhang, Lu, Wang, Chen, and Wang]{zeng2023conzic}
Zequn Zeng, Hao Zhang, Ruiying Lu, Dongsheng Wang, Bo~Chen, and Zhengjue Wang.
\newblock Conzic: Controllable zero-shot image captioning by sampling-based polishing.
\newblock In \emph{Proceedings of the IEEE/CVF conference on computer vision and pattern recognition}, pp.\  23465--23476, 2023.

\bibitem[Zhan et~al.(2024)Zhan, Lin, Wang, Wang, and Wu]{zhan2024medm2g}
Chenlu Zhan, Yu~Lin, Gaoang Wang, Hongwei Wang, and Jian Wu.
\newblock Medm2g: Unifying medical multi-modal generation via cross-guided diffusion with visual invariant.
\newblock In \emph{Proceedings of the IEEE/CVF conference on computer vision and pattern recognition}, pp.\  11502--11512, 2024.

\bibitem[Zhang et~al.(2024{\natexlab{a}})Zhang, Yang, Yu, Fan, Jiang, Huang, and Han]{zhang2024attribute}
Ke~Zhang, Yan Yang, Jun Yu, Jianping Fan, Hanliang Jiang, Qingming Huang, and Weidong Han.
\newblock Attribute prototype-guided iterative scene graph for explainable radiology report generation.
\newblock \emph{IEEE Transactions on Medical Imaging}, 2024{\natexlab{a}}.

\bibitem[Zhang et~al.()Zhang, Rao, and Agrawala]{zhang2023adding}
Lvmin Zhang, Anyi Rao, and Maneesh Agrawala.
\newblock Adding conditional control to text-to-image diffusion models.

\bibitem[Zhang et~al.(2020)Zhang, Wang, Xu, Yu, Yuille, and Xu]{zhang2020radiology}
Yixiao Zhang, Xiaosong Wang, Ziyue Xu, Qihang Yu, Alan Yuille, and Daguang Xu.
\newblock When radiology report generation meets knowledge graph.
\newblock In \emph{Proceedings of the AAAI conference on artificial intelligence}, volume~34, pp.\  12910--12917, 2020.

\bibitem[Zhang et~al.(2024{\natexlab{b}})Zhang, Yao, Wang, Jha, Durak, Keles, Medetalibeyoglu, and Bagci]{zhang2024diffboost}
Zheyuan Zhang, Lanhong Yao, Bin Wang, Debesh Jha, Gorkem Durak, Elif Keles, Alpay Medetalibeyoglu, and Ulas Bagci.
\newblock Diffboost: Enhancing medical image segmentation via text-guided diffusion model.
\newblock \emph{IEEE Transactions on Medical Imaging}, 2024{\natexlab{b}}.

\bibitem[Zhang et~al.(2018)Zhang, Yang, and Zheng]{zhang2018translating}
Zizhao Zhang, Lin Yang, and Yefeng Zheng.
\newblock Translating and segmenting multimodal medical volumes with cycle-and shape-consistency generative adversarial network.
\newblock In \emph{Proceedings of the IEEE conference on computer vision and pattern Recognition}, pp.\  9242--9251, 2018.

\end{thebibliography}
\bibliographystyle{iclr2026_conference}

\appendix

\section*{Technical Appendices}

\section{Related Work}\label{sec_related}

\paragraph{Medical Image Generation.}
Owing to data scarcity, medical image generation is gaining increasing significance as a research focus. 
Early studies~\citep{chartsias2017adversarial,guo2020anatomic,huo2018synseg,zhang2018translating} employed GANs to augment medical imaging data, improving the performance of downstream medical tasks. 
In recent years, diffusion models~\citep{ho2020denoising,rombach2022high} have shown stable training and promising performance in high-quality image generation tasks, motivating a growing body of work to explore their application in medical image generation~\citep{mao2025medsegfactory,lyu2022conversion,chambon2022roentgen,gungor2023adaptive}. 
Fast-DDPM~\citep{jiang2025fast} and LLCM~\citep{polamreddy2025leapfrog} focus on accelerating the diffusion-based generation of medical images. 
EMIT-Diff~\citep{zhang2024diffboost} and MAISI~\citep{guo2025maisi} incorporate ControlNet~\citep{zhang2023adding} to achieve anatomically or semantically controllable medical image synthesis. 
By introducing non-linear Kolmogorov-Arnold Networks (KANs)~\citep{liu2024kan} and globally modeling Transformers~\citep{vaswani2017attention,Peebles2022DiT} as diffusion backbones, U-KAN~\citep{li2025u} and Med-Art~\citep{guo2025med} demonstrate state-of-the-art performance in generating medical images across diverse modalities, which motivates the consideration of backbone design for discrete diffusion models in MeDiM.

\paragraph{Medical Report Generation.}
Medical report generation is commonly facilitated by VLMs~\citep{li2022blip,li2023blip,zeng2023conzic} endowed with domain-specific medical knowledge. 
\citet{yang2022knowledge} enhanced the medical literacy of VLMs by combining general medical background with image-specific clinical knowledge, while \citet{zhang2020radiology} designed a medical knowledge graph to further enrich medical VLMs. 
R2Gen~\citep{chen-emnlp-2020-r2gen} and R2GenCMN~\citep{chen-acl-2021-r2gencmn} generate radiology reports based on memory-driven Transformer architectures. 
\citet{chen2023fine} improved the interpretability of medical report generation by associating local medical image regions with medical terminology in the report, while \citet{liu2024factual} enhanced image–report alignment through fact-guided contrastive learning. 
RGRG~\citep{tanida2023interactive} employs region-level guidance for medical report generation, while \citet{zhang2024attribute} addresses its limitation of neglecting shared attributes for each local region via attribute prototype guidance.
\citet{liu2024bootstrapping} leveraged priors from LLMs to improve medical report generation. 
MLRG~\citep{liu2025enhanced} enhances medical visual information by integrating multiple views of the same medical image. 
However, most existing approaches are limited to specific medical report generation tasks. 
To overcome this issue, we present MeDiM, which formulates the joint discrete distribution of medical images and reports independent of imaging modalities and organ-specific characteristics.

\paragraph{Unified Multimodal Understanding and Generation.}
Unified multimodal models aim to build an MLLM, modeling and reasoning over both image and textual sequences. 
Early studies~\citep{ge2307planting,ge2024seed,sun2024generative,sun2023emu,li2022blip} attempted to combine CLIP~\citep{radford2021learning}, which aligns the visual–language space, with LLMs to process images and text separately. 
Given that CLIP operates in a continuous visual space, some studies~\citep{liu2024world,team2024chameleon,wu2024vila} have explored using VQ-VAE~\citep{van2017neural} to represent visual information as discrete sequences. 
Meanwhile, numerous studies~\citep{yu2023scaling,lu2024unified,xie2024show,wu2025janus} discussed leveraging the strengths of both encoders. 
Studies such as LWM~\citep{liu2024world} and Chameleon~\citep{team2024chameleon} have shown that discrete visual features can be integrated with language tokens into a unified sequence, facilitating joint cross-modal modeling. 
Although this design obviates the need for modality-specific components, it incurs substantial training costs. 
Liquid~\citep{wu2024liquid} introduces a pre-trained LLM as the backbone to solve this issue. 
However, AR suffers from low inference efficiency.
Given the efficiency in generation and success in understanding of discrete diffusion models, UniD3~\citep{hu2022unified} and UniDisc~\citep{swerdlow2025unified} adopted them as a unified framework. 
MMaDA~\citep{yang2025mmada} further employs LLaDA~\citep{nie2025large} as the backbone for a unified discrete diffusion model. 
However, MMaDA is restricted to a single diffusion-based MLLM backbone, whereas MeDiM can leverage the wider spectrum of AR MLLM backbones, providing greater scalability.

\begin{table*}[ht]
\centering
\caption{Ablation study for MeDiM on the backbone and components, evaluated with mean medical report understanding metrics (mB-1, mB-2, mB-3, mMTR, mR-L) and mean image generation metrics (mFID, mIS) on MIMIC-CXR and PathGen.}
\label{tab_abla}
\small
\resizebox{\textwidth}{!}{
\begin{tabular}{c ccccc cc} % 去掉前两列后的竖线
\toprule
\multirow{2}{*}{Settings} 
& \multicolumn{5}{c}{Report Generation} 
& \multicolumn{2}{c}{Image Generation} \\
\cmidrule(lr){2-6} \cmidrule(lr){7-8}
& mB-1 $\uparrow$ & mB-2 $\uparrow$ & mB-3 $\uparrow$ & mMTR $\uparrow$ & mR-L $\uparrow$ & mFID $\downarrow$ & mIS $\uparrow$ \\
\midrule
MeDiM & 0.256 & 0.134 & 0.073 & 0.262 &  0.261 & 20.40 & 3.57 \\
\midrule
w/  DiT backbone & 0.195 & 0.091 & 0.040 & 0.214 &0.200 & 63.22 & 2.81 \\
w/  UniDisc backbone & 0.223 & 0.098 & 0.051 & 0.255 & 0.217 & 51.59 & 3.05\\
w/o pretrained MLLM weight & 0.205 & 0.092 & 0.044 & 0.229 & 0.212 & 68.27 & 2.83 \\
\midrule
w/o timestep embedding & 0.221 & 0.107 & 0.049 & 0.246 & 0.238 & 40.03 & 3.13\\
w/o  AdaLN designs & 0.232 & 0.108 & 0.056 & 0.247 & 0.249 & 32.68 & 3.16  \\
w/  causal mask & 0.152 & 0.068 & 0.025 & 0.142 & 0.179 & 143.72 & 2.02 \\
\bottomrule
\end{tabular}
}
\end{table*}

\section{Ablation Study}\label{sec_abla}
We investigate the contribution of the MLLM backbone and its associated architectural components to the overall performance of MeDiM.
This analysis allows us to disentangle the impact of backbone selection and design choices.

\begin{figure*}[!t]
    \centering
    \setlength{\abovecaptionskip}{-1mm} %调整caption与图的距离
    \setlength{\belowcaptionskip}{-5mm}%调整caption与下文的距离
    \includegraphics[width=1\linewidth]{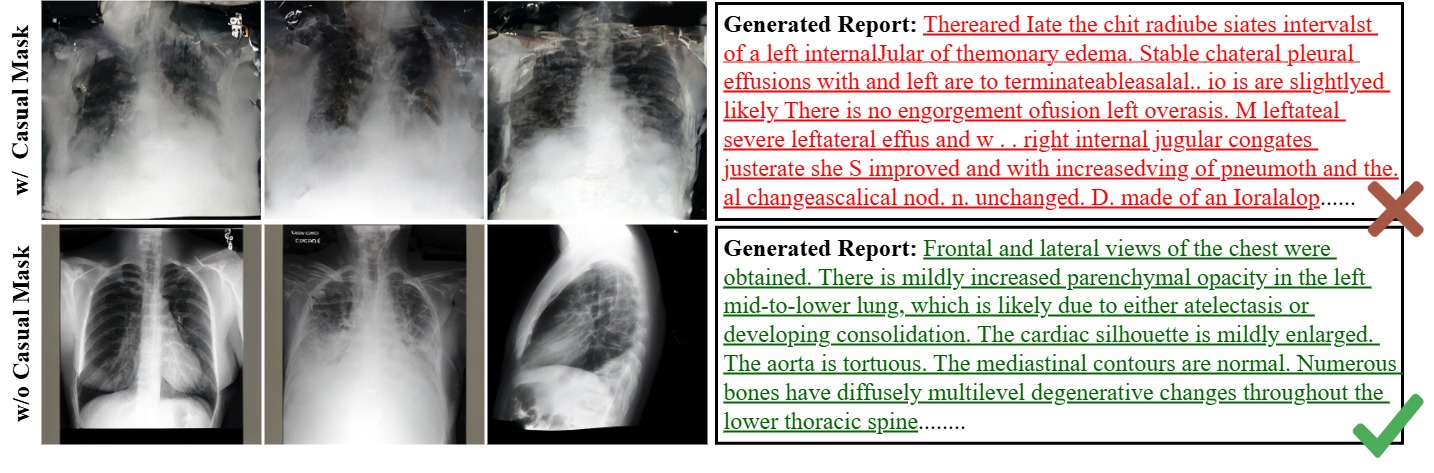}
     \caption{Visual results from the ablation study evaluating the effect of pre-trained MLLM weights.}
     \label{fig_abla}
\end{figure*}

\paragraph{Backbone.}\label{sec_backbone}
We use DiT~\citep{Peebles2022DiT} and a pretrained UniDisc backbone as backbone baselines to examine the impact of backbone choice.
We also assess the contribution of the pretrained distributional alignment prior in MLLM to MeDiM’s performance. 
As shown in Tab.~\ref{tab_abla}, replacing the MLLM backbone with different backbones or removing the pretrained MLLM weights both lead to noticeable degradation for MeDiM's performance.
%This indicates that MLLMs endowed with alignment priors are more suitable as backbones for unified discrete models in medical generation and understanding.

\paragraph{Components.}
We further conduct ablation studies on three architectural components of MeDiM to assess their contributions: timestep embeddings, AdaLN designs, and causal mask removal. 
We remove the timestep embeddings and replace the AdaLN designs with a standard LayerNorm, thereby eliminating temporal conditioning.
For the AdaLN ablation, we retain timestep embeddings but inject them directly into the token representations through weighted addition, without AdaLN components. 
The ablation results in Tab.~\ref{tab_abla} demonstrate that timestep embeddings and AdaLN provide essential temporal conditioning and effective feature modulation. 
In contrast, enforcing a causal mask severely disrupts multimodal alignment, resulting in blurry image boundaries and semantically incoherent reports (see Fig.~\ref{fig_abla}). 

\section{Limitation}
Although MeDiM achieves promising results on unified medical multimodal generation tasks, it has not yet exceeded or matched expert medical models across all metrics at evaluation, with shortfalls confined to a limited subset.
We plan to improve MeDiM by incorporating MLLM backbones with medical domain background knowledge, aiming to bridge the gap with SoTA medical expert models.

\end{document}